\pdfoutput=1

\documentclass[11pt]{article}

\usepackage{acl}
\usepackage{amsmath}
\usepackage{algpseudocode}
\usepackage{algorithm}
\usepackage{amssymb}
\usepackage{booktabs}
\usepackage{times}
\usepackage{latexsym}

\usepackage[T1]{fontenc}

\usepackage[utf8]{inputenc}

\usepackage{microtype}

\usepackage{inconsolata}
\usepackage{xcolor}
\usepackage{xspace}
\usepackage{multirow}
\usepackage{verbatim}
\usepackage{amsfonts}
\usepackage{graphicx}
\usepackage{array, makecell}
\usepackage{subfigure}
\usepackage{soul}

\newcommand{\draftcomment}[1]{#1}

\definecolor{class_pink}{HTML}{e78ac3}
\definecolor{class_green}{HTML}{a6d854}
\definecolor{class_blue}{HTML}{8da0cb}
\definecolor{class_teal}{HTML}{66c2a5}
\definecolor{class_orange}{HTML}{fc8d62}
\definecolor{class_ood}{HTML}{b3b3b3}

\newcommand{\hlc}[2][yellow]{ {\sethlcolor{#1} \hl{#2}} }
\renewcommand{\draftcomment}[1]{}

\title{Out-of-Distribution Detection through Soft Clustering with \\ Non-Negative Kernel Regression}

\newcommand{\aspace}{\hspace{1em}}
\newcommand{\usc}{$^{\heartsuit}$}

\newcommand{\spain}{$^{\ddagger}$}
\newcommand{\industry}{$^{\clubsuit}$}
\author{
    Aryan Gulati\usc\aspace 
    Xingjian Dong\usc\aspace 
    Carlos Hurtado\spain \aspace 
    Sarath Shekkizhar\industry\aspace \\
    \textbf{Swabha Swayamdipta}\usc \aspace 
    \textbf{Antonio Ortega}\usc\aspace \\
    \usc University of Southern California, Los Angeles, USA \aspace \industry Tenyx \\
    \spain Universitat Politecnica de Catalunya, Barcelona, Spain \\
    \texttt{\{aryangul,xdong404\}@usc.edu}
}

\begin{document}
\maketitle

\begin{abstract}
As language models become more general purpose, increased attention needs to be paid to detecting out-of-distribution (OOD) instances, i.e., those not belonging to any of the distributions seen during training. 
Existing methods for detecting OOD data are computationally complex and storage-intensive.
We propose a novel soft clustering approach for OOD detection based on non-negative kernel regression.
Our approach greatly reduces computational and space complexities (up to $11\times $ improvement in inference time and 87\% reduction in storage requirements) and outperforms existing approaches by up to 4 AUROC points on four different benchmarks.
We also introduce an entropy-constrained version of our algorithm, which leads to further reductions in storage requirements (up to 97\% lower than comparable approaches) while retaining competitive performance.
Our soft clustering approach for OOD detection highlights its potential for detecting tail-end phenomena in extreme-scale data settings.
\end{abstract}

\section{Introduction}
Despite the successes of generalized models of natural language, the challenge of generalization to out-of-distribution~(OOD) data---data that differs from the training data distribution--- remains \citep{elsahar-galle-2019-annotate,liu-etal-2024-good-llms}. 
This can be a limiting obstacle in known, sensitive domains like medicine and finance \citep{yang-etal-2023-distribution,salehi2022a}, or even in ``domains'' which are unknown or imperceptible to humans \cite{plank2016nonstandard}.
OOD shifts are also important in detecting long tail phenomena \cite{lewis-etal-2021-question,liu-etal-2022-challenges}, which are critical to ensure robust and reliable application of modern language models.

\begin{figure}[H]
    \centering
    \includegraphics[width=\linewidth]{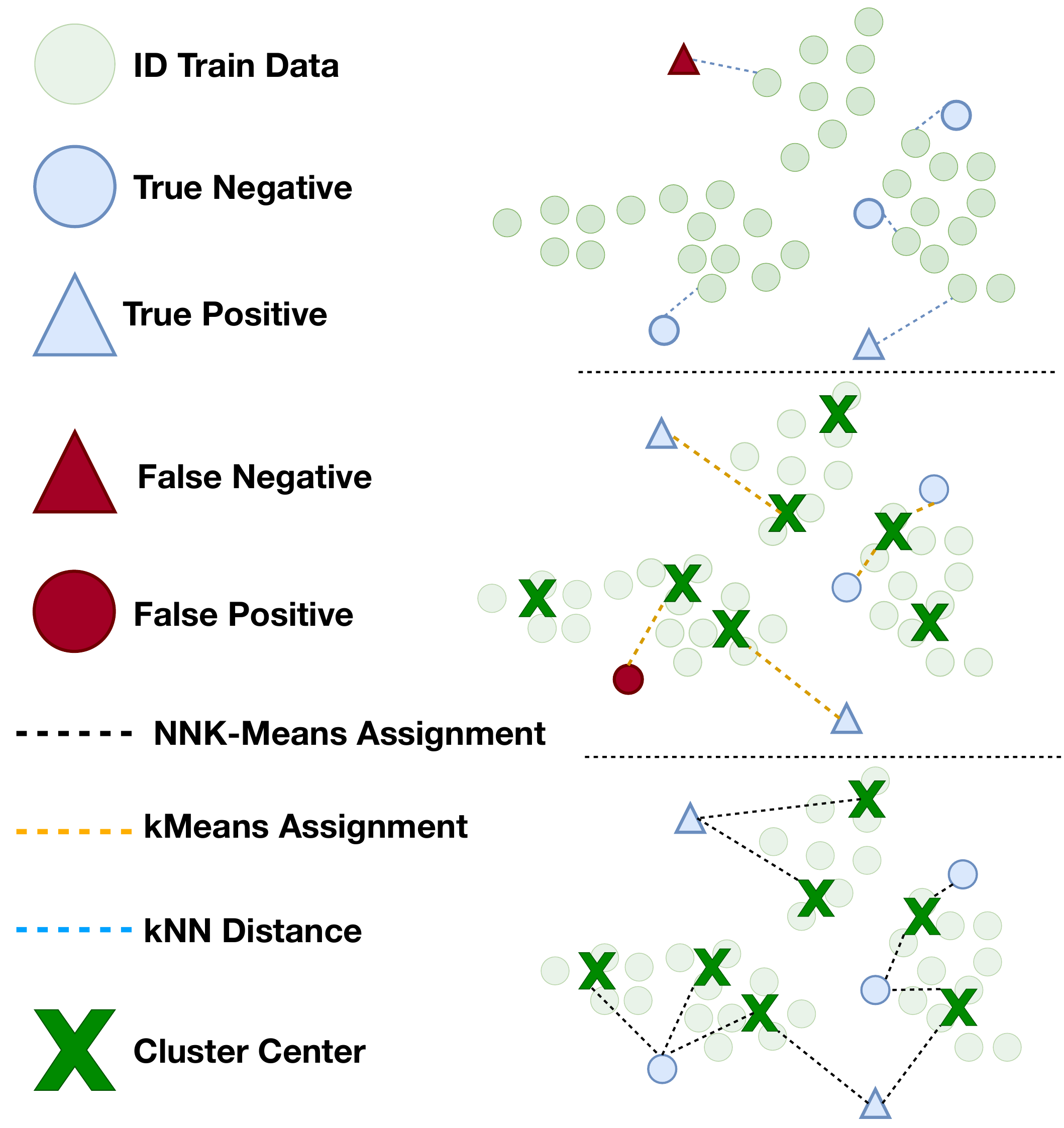}
    \caption{Illustration comparing KNN (top) with kMeans (middle) and our proposed NNK-Means (bottom). The use of soft-clustering allows our method to detect OOD instances even when they are close to ID training data. It also better captures the underlying data geometry, enabling more accurate identification of ID data points than kMeans.} 
    \label{fig:nnk_means_polytope}
\end{figure}

While OOD detection has been extensively studied (\S\ref{sec:related}), most approaches have limitations preventing them from being applied broadly. 
Existing distance-based approaches for OOD detection \cite{Sun2022ICMLDeepNN,10.1145/335191.335388,10.1007/978-3-642-01307-2_86} are often not scalable as they rely on storing the entire in-distribution (ID) training set. This is particularly challenging given the size of training data for LLMs.
Approaches that improve scalability make strong assumptions about the distribution of data (e.g., the ID data does not have small clusters \cite{he2003discovering}) or are applicable only when the data is labeled \cite{Lee-etal-mahalanobis}.

While requiring lower storage and computation, classification-based approaches for OOD detection are typically limited to cases where labeled data is available \cite{hendrycks2017a}.
Moreover, they perform worse than distance-based approaches \cite{liang2018enhancing}.   

In this work, we present a clustering approach for OOD detection that (i) makes no assumptions about the underlying data distribution, (ii) applies to both labeled and unlabeled data, (iii) is scalable, and (iv) is compute and storage-efficient.
Our OOD detection method builds on a dictionary-based approach that leverages a non-negative kernel regression (NNK)-based soft clustering technique called NNK-Means \citep{shekkizhar2022nnk} (see \autoref{fig:nnk_means_polytope}). 
Soft clustering, i.e., associating each sample with multiple cluster centers in the data manifold, leads to a better approximation of the ID data and, consequently, improved OOD detection. 
It also requires fewer clusters and is therefore storage-efficient.
We are the first to leverage soft clustering for OOD detection with text.
Moreover, to avoid dependence on the number of cluster centers---the critical limitation in most clustering algorithms---we introduce a new, improved formulation of NNK-Means, proposing an entropy-constrained data-driven selection process \footnote{Our code is available at \url{https://github.com/STAC-USC/NNK-Means-OOD}}.

We empirically validate the performance of NNK-Means for OOD detection on 4 benchmark datasets.
We show that it consistently achieves superior or comparable performance relative to state-of-the-art approaches \citep{liu2020energy,Sun2022ICMLDeepNN} while requiring over an order of magnitude lower storage and inference time.
We also find that our approach is applicable across a variety of settings, effectively leveraging ID labels when they are present but providing competitive performance when they are not, and maintains high performance when using different types of embeddings.
Overall, we find that our soft-clustering based approach yields state-of-the-art OOD detection performance, while improving memory and computational efficiency - particularly when using our improved formulation with entropy constraints.

\section{Related Work}
\label{sec:related}

OOD detection methods in NLP broadly fall into two categories: (i) post-hoc methods that detect OOD instances after deriving their representations from pre-trained language models (PLMs) and (ii) works focused on learning representations that improve OOD detection. 

\paragraph{Post-hoc OOD Detection} These methods are typically applied to the dataset representations, which can either be \textit{Pre-trained Representations} obtained directly from PLMs or \textit{Fine-tuned Representations} obtained after fine-tuning the PLMs for a particular task. 
Post-hoc methods can be further divided into two categories. First, \textbf{distance-based methods} compute the minimum distance to new data from ID training data as the OOD score. For example, \citet{Lee-etal-mahalanobis} computes the class-wise Mahalanobis distance between class centroids and a query point to obtain an OOD score. \citet{xu2020deep} proposed Gaussian Discriminant Analysis (GDA), which leverages Euclidean and Mahalanobis distances with generative classifiers to identify OOD instances. \citet{Sun2022ICMLDeepNN} directly uses the distance to the $k$th nearest neighbor (KNN). However, these approaches require storing the entire ID training set, significantly increasing memory requirements. Alternatively, based on the intuition that a classifier output distribution tends to reflect training distribution, \textbf{classifier-based methods} leverage the output logits to get a confidence score for OOD detection. 
The most frequently used and simple such method uses the Maximum Softmax Probability (MSP) of the classifier as confidence, as introduced by \citet{hendrycks2017a} and later improved by ODIN \cite{liang2017enhancing} by adding temperature scaling and input preprocessing. To tackle the over-confidence problem of MSP, \citet{liu2020energy} introduces Energy, an energy-based scoring function to better detect OOD data. \citet{yilmaz2022d2u} instead proposes Distance-to-Uniform (D2U) to find the OOD data whose output distribution is closer to a uniform distribution.

\paragraph{Learning Representations for OOD Detection} Many methods employ Supervised or Margin-based Contrastive Loss \cite{zhou2021contrastive} for OOD detection, which increases the similarity of instance pairs if they belong to the same class and decreases it otherwise. Various variants have introduced multiple improvements to enhance discrimination performance, such as Adversarial Contrastive Learning \cite{zeng2021adversarial}, KNN-enhanced Contrastive Learning (KNN-CL) \cite{zhou2022knn}, and Reassigned Contrastive Learning (RCL) \cite{wu2022revisit}. Apart from Contrastive Learning, \citet{xu2021unsupervised} utilizes features from all layers of PLMs to form Mahalanobis Distance Features (MDF), and GNOME \cite{chen2023fine} combines MDF from both pre-trained and fine-tuned models, while Avg-avg \cite{chen2022holistic} simply averages all token representations in each intermediate layer to form the sentence representation for OOD detection. 

Additionally, obtaining OOD data in real-world scenarios is challenging; thus, many methods use pseudo-OOD data for representation learning \cite{zhan2021out, shu2021odist, lang2022estimating, xu2022contrastive, kim2023pseudo}.
Besides these, methods like DATE \cite{manolache2021date}, PTO \cite{ouyang2023prefix}, and BLOOD \cite{jelenic2023out} do not fit into these categories but have also achieved notable results.

Our work is a \textbf{post-hoc} method, which focuses primarily on techniques to detect OOD samples irrespective of the representations used. 
Our proposed method is computationally efficient, providing the memory benefits of clustering and classifier-based techniques while performing comparably with distance-based methods.

\section{NNK-Means and Variants}
\label{sec:method}

We briefly present the background on soft clustering via NNK-Means \citep{shekkizhar2022nnk} for modeling a data distribution (\S\ref{sec:pre}).
Next, we present our extension of the method via the introduction of an entropy constraint (\S\ref{sec:ecnnk}).

\subsection{Background}
\label{sec:pre}

Conventional clustering methods, such as kMeans \cite{he2003discovering}, are trained in two steps: (i) \textit{coding}: each training item is assigned to \textit{one} existing cluster (corresponding to the nearest cluster center), and (ii) \textit{dictionary update}: new cluster centers are computed, where each cluster center (dictionary atom) is the average of all training items assigned to the cluster (see \autoref{fig:nnk_means_polytope}, middle).

In contrast, a soft-clustering approach such as NNK-Means operates as follows.
(i) Coding: each training item is assigned to \textit{multiple} cluster centers (\textbf{sparse coding}), with non-negative weights that quantify similarity to the cluster center (larger weights for higher similarity between input and cluster center). This soft clustering allows more flexible representations with lower storage (fewer clusters can represent the data). 
(ii) Dictionary Update: the new cluster centers (\textbf{atoms}) are obtained as weighted averages of the inputs assigned to the cluster, where the weights are non-negative. 
The set of cluster centers is designed to minimize reconstruction error on the training data.
\autoref{fig:nnk_means_polytope} (bottom) illustrates this approach.

Formally, given a dataset of $N$ data points represented by a matrix $\boldsymbol{X} \in \mathbb{R}^{d \times N}$, the goal is to learn a dictionary matrix $\boldsymbol{D} \in \mathbb{R}^{d \times M}$ (where each column represents a cluster center) and a sparse weight matrix $\boldsymbol{W} \in \mathbb{R}^{M \times N}$ which generates sparse linear combinations of the columns of $\boldsymbol{D}$ that approximate the training data:
\begin{align}
\hat{\boldsymbol{D}}, \hat{\boldsymbol{W}} &= \underset{\substack{\boldsymbol{D}, \boldsymbol{W} : \forall i, \boldsymbol{w}_i \geq 0, \\ \left \|\boldsymbol{w}_i\right\|_0 \leq k}}{\arg\min} \| \boldsymbol{X} - \boldsymbol{DW}\|^2_2 \label{eq:nnkm_dw}
\end{align}
Here, each column of $\boldsymbol{W}$,  $\boldsymbol{w}_i$, is sparse, with at most $k$ non-zero entries. 
To achieve this, NNK-Means alternates between \textit{sparse coding} and \textit{dictionary/cluster update} as follows, until a convergence criterion is reached. 

\paragraph{Sparse Coding} We find a $\boldsymbol{W}$ that minimizes reconstruction error with the current dictionary. 
We can rewrite the objective in \eqref{eq:nnkm_dw} to instead use a kernelized representation of the input data $\boldsymbol{\Phi} = \phi(\boldsymbol{X}) \in \mathbb{R}^{N \times N}$. Since each atom is a nonnegative linear combination of  elements of $\boldsymbol{\Phi}$, 
the dictionary matrix can be written $\boldsymbol{D} = \boldsymbol{\Phi}\boldsymbol{A} \in \mathbb{R}^{d \times M}$, where 
$\boldsymbol{A} \in \mathbb{R}^{N \times M}$ is the dictionary coefficients matrix containing the weights.
Then, we can kernelize the minimization objective from \eqref{eq:nnkm_dw} and find each column of $\boldsymbol{\hat{W}}$ as 
\begin{align}
\hat{\boldsymbol{w}_i}=\underset{\boldsymbol{w}_i \geq 0,\left\|\boldsymbol{w}_i\right\|_0 \leq k}{\arg\min } 
\left\|\boldsymbol{\phi}_i-\boldsymbol{\Phi} \boldsymbol{A} \boldsymbol{w}_i\right\|_2^2
, \label{eq:nnkm_w}
\end{align}
where $\boldsymbol{\phi}_i$ corresponds to the kernel representation of data $\boldsymbol{x}_i$.
Finding $\hat{\boldsymbol{w}_i}$ from \eqref{eq:nnkm_w} involves handling an $N \times N$ kernel matrix, resulting in run times that would scale poorly with the dataset size. 
\citeposs{shekkizhar2020graph} geometric insight into the NNK objective enables the efficient computation of each $\hat{\boldsymbol{w}_i}$ from a small subset of the data, specifically the $k$-nearest neighbors of each point.
Thus, \eqref{eq:nnkm_w} can be rewritten for each data point and solved with NNK as 
\begin{align}
\label{eq:opt-nnk}
\hat{\boldsymbol{w}}_{i, S}=\underset{\boldsymbol{\theta}_i \geq 0}{\arg \min }\left\|\boldsymbol{\phi}_i-\boldsymbol{\Phi} \boldsymbol{A}_S \boldsymbol{\theta}_i\right\|_2^2 \text { and } \hat{\boldsymbol{w}}_{i, S^{c}}=\mathbf{0}
,
\end{align}
where the set $S$ corresponds to a subset of the dictionary atoms $\boldsymbol{\Phi} \boldsymbol{A}$ that can have nonzero influence. 
The resulting sparse coefficients have a geometric interpretation, such that the sparse set of selected atoms forms a convex polytope around each point in the dataset \cite{shekkizhar2020graph}.

\paragraph{Dictionary Update}
Given the sparse codes $\boldsymbol{W}$ computed in the first step, this second step updates the dictionary coefficients matrix $\boldsymbol{A}$ to minimize the reconstruction error: 
\begin{equation}\label{dict_update}
\boldsymbol{A} = \boldsymbol{W}^{\top}(\boldsymbol{W}\boldsymbol{W}^{\top})^{-1}. 
\end{equation}
This update rule is similar to the Method of Optimal Directions \citep{engan1999method} and has the advantage of keeping the cluster centers in the same space as input data.

A limitation of NNK-Means is that the number of atoms in the dictionary, $M$, is a hyperparameter. 
While dictionaries with a larger set of atoms can improve representation, 
they increase the complexity of coefficient selection, 
while also requiring more storage.  
In NNK-Means, there is no obvious way to adjust the number of atoms other than training the system with a new choice of $M$. 

\subsection{Entropy-Constrained NNK-Means}
\label{sec:ecnnk}

To address these limitations, we propose \textbf{E}ntropy-\textbf{C}onstrained \textbf{NNK-Means} (\textbf{EC-NNK-Means}). Our new approach estimates the number of points that select each cluster from the sparse coding weights in $\boldsymbol{W}$. The percentage of points selecting a cluster can be viewed as ``cluster probability,'' which quantifies the importance of the cluster. 
Then, we introduce an entropy-based regularization term into the cluster optimization, which favors selecting atoms representing more data points (i.e., higher probability/lower entropy atoms). 

Consider a query $\boldsymbol{q} = \boldsymbol{x}_i$ and the set $S$ of its $k$-nearest dictionary atoms. 
We can expand the minimization objective in \eqref{eq:opt-nnk} 
\begin{equation}\label{nnk_opt}
    \boldsymbol{\theta}_i=\underset{\boldsymbol{\theta} \geq 0}{\arg \min } \frac{1}{2} \boldsymbol{\theta}^{\top} \boldsymbol{K}_{S, S} \boldsymbol{\theta}-\boldsymbol{\theta}^{\top} \boldsymbol{K}_{S, \boldsymbol{q}}
    ,
\end{equation}
where $\boldsymbol{K}_{Y, Z} = \phi(Y)^{\top}\phi(Z)$ is the chosen kernel function that encodes similarity between any given sets of vectors $Y$ and $Z$. 

In \eqref{nnk_opt}, cluster assignments are influenced by the similarities between the query and its nearest cluster centers ($\boldsymbol{K}_{S, \boldsymbol{q}}$) and between cluster centers ($\boldsymbol{K}_{S, S}$).
This results in each point being assigned to a non-redundant set of its most similar atoms but does not account for the size of each cluster. 
The NNK-Means assignment objective can be modified to consider also the probability that a given point belongs to each cluster, represented by $\boldsymbol{p} \in \mathbb{R}^M$.
To do this, we include an entropy regularization term that penalizes the least selected (lower probability/higher entropy) clusters: 
\begin{equation}\label{entropy_opt}
    \boldsymbol{\theta}_i=\underset{\boldsymbol{\theta} \geq 0}{\arg \min } \frac{1}{2} \boldsymbol{\theta}^{\top} \boldsymbol{K}_{S, S} \boldsymbol{\theta}-\boldsymbol{\theta}^{\top} \boldsymbol{K}_{S, \boldsymbol{q}}+\lambda \boldsymbol{\theta}^{\top} \log\boldsymbol{p}_S
    ,
\end{equation}
where $\boldsymbol{p}_S$ corresponds to the probability of each atom in the set $S$, and $\lambda$ is a hyperparameter that controls the relative influence between the kernel similarity and probability. 

The probability $\boldsymbol{p}_i$ of atom $i$ being chosen is determined by:
\begin{equation}\label{p_i1}
p_i = \frac{\sum_j \mathbb{I}(\boldsymbol{W}_{i,j} > 0)}{\sum_i \sum_j \mathbb{I}(\boldsymbol{W}_{i,j} > 0)}
\end{equation}
where $\mathbb{I}(\cdot)$ is an indicator function that is equal to $1$ if the condition inside is true. This probability is defined as the number of data points assigned to atom $i$ data with a non-zero weight normalized over the size of the dataset.

The additional entropy term added to the NNK-Means objective ($\boldsymbol{\theta}^{\top} \log\boldsymbol{p}_S$) can also be regarded as the cross-entropy between the new sparse code $\boldsymbol{\theta}$ and the current $\log \boldsymbol{p}_S$. Minimizing this term leads to an assignment that aligns both distributions as closely as possible. Consequently, atoms that are assigned more elements during training have a higher probability of being selected by a new data point, while the reverse is true for atoms having less data assigned during training.

To adaptively learn a dictionary of a size appropriate to the data, we iteratively prune the set of $M$ dictionary atoms to a final dictionary of size $\hat{M}$. 
Atoms with a lower probability will have fewer data points assigned in future weight assignments and eventually, their corresponding $p_i$ will reach $0$ and they will be removed from the dictionary. This process allows for the selection of a larger initial number of atoms than the original NNK-Means, enhancing the likelihood of choosing atoms that are representative of the underlying data, while also improving efficiency by eliminating unimportant atoms.
The full training procedure is described in Algorithm \ref{alg:ecnnkm}. 

\begin{algorithm}[ht]
\caption{Entropy-Constrained NNK-Means}
\label{alg:ecnnkm}
\renewcommand{\arraystretch}{0.8}
\textbf{Input:} Dataset $\boldsymbol{X}$, training steps $I$, dictionary initial size $N$
\begin{algorithmic}[1]
\State $\boldsymbol{A} = \{\text {Dictionary initialized with kMeans++}\}$
\State $\boldsymbol{p} = \left[ \frac{1}{N}, \frac{1}{N}, \ldots, \frac{1}{N} \right]_{1 \times N}$
\For {iter in $I$}
\State $\boldsymbol{W} = \text{EC-NNK sparse codes}$
\State $p_i = \frac{\sum_j \mathbb{I}(\boldsymbol{W}_{i,j} > 0)}{\sum_i \sum_j \mathbb{I}(\boldsymbol{W}_{i,j} > 0)}$
\State $\boldsymbol{A} = \boldsymbol{W}^{\top}(\boldsymbol{W}\boldsymbol{W}^{\top})^{-1}$
\State $\boldsymbol{\hat{A}} = \boldsymbol{A} \setminus \boldsymbol{A}_i, \: \forall i : p_i = 0$
\EndFor
\end{algorithmic}
\textbf{Output:} Dictionary $\boldsymbol{\hat{A}}$ of size $\hat{N}$
\end{algorithm}

\section{NNK-Means for OOD Detection}

In this section, we formally formulate the OOD detection problem (\S\ref{sec:preliminaries}) and describe how to use NNK-Means for OOD detection (\S\ref{sec:anomaly}).

\subsection{OOD Detection Formulation}
\label{sec:preliminaries}

Define an in-distribution (ID) training dataset $D_{\text{ID}} = \{(\boldsymbol{x}_1, y_1), (\boldsymbol{x}_2, y_2), \hdots, (\boldsymbol{x}_N, y_N)\}$ where $\boldsymbol{x}_i$ is a text entry and $y_i \in \{1, \hdots, C\}$ is the corresponding label. 
We also assume access to an encoder $E: \boldsymbol{x} \rightarrow \mathbb{R}^d$ that maps the text to a $d$-dimensional feature space. 
We formulate our OOD Detection problem as a binary classification task to determine whether or not a sample is OOD with respect to the training distribution, $D_{\text{ID}}$, following prior work \cite{liu2020energy, xu2021unsupervised, chen2023fine}.
The goal is to generate an OOD score $O(\boldsymbol{x}; E)$ which represents the probability of an instance being out-of-distribution, 
and the final decision $G_\epsilon(\boldsymbol{x}; E)$ can be made by:
\begin{equation}
G_\epsilon(\boldsymbol{x}; E) = 
\begin{cases} 
\text{ID} & \text{if } O(\boldsymbol{x}; E) \geq \epsilon \\
\text{OOD} & \text{if } O(\boldsymbol{x}; E) < \epsilon 
\end{cases},
\end{equation}
where $\epsilon$ represents a chosen threshold. In practice, the threshold is chosen to ensure about 95\% recall.

\paragraph{Pipeline} Our pipeline is as follows: for a given sample $\boldsymbol{x}$, we first obtain its representations using the encoder $E$. 
These representations $E(\boldsymbol{x})$ are then passed to OOD detection methods, which can be either classifier-based or post-hoc (described in \S\ref{sec:related}), finally yielding an OOD score, $O(\cdot)$. 
The backbone model of encoder $E$ can be a PLM or a fine-tuned version $E'$, which is trained on a classification task using the ID training data.

\subsection{Generating OOD Scores with NNK-Means}
\label{sec:anomaly}

The dictionary and assignments learned by NNK-Means are optimized to minimize the reconstruction error of the training data. 
New data that cannot be properly reconstructed using this dictionary, i.e., data with a higher reconstruction error, is more likely to be out-of-distribution.
Therefore, we can use the definition of reconstruction error from \eqref{nnk_opt} as an OOD score. 
For any query $\boldsymbol{q} \in \mathbb{R}^d$, we define its OOD score $O(\boldsymbol{q})$ as
\begin{equation}\label{eq:reconstruction-error}
    O(\boldsymbol{q}) = \frac{1}{2} \boldsymbol{\theta}_S^{\top} \boldsymbol{K}_{S, S} \boldsymbol{\theta}_S-\boldsymbol{\theta}_S^{\top} \boldsymbol{K}_{S, \boldsymbol{q}}
\end{equation}
Note that the value of $\boldsymbol{\theta}$ is obtained by minimizing the objective in \eqref{nnk_opt}, and $S$ represents the set of $k$-nearest dictionary atoms to $\boldsymbol{q}$.

We also propose C-NNK-Means, a class-wise extension incorporating label information when labeled ID data is available.
Here, rather than learning one dictionary $\boldsymbol{D}$ for the entire ID dataset, we learn a separate dictionary $\boldsymbol{D}_c$ for each ID class. 
Then, the OOD score is:
\begin{equation} \label{eq:c-error}
O_c(\boldsymbol{q}) = \min_c \frac{1}{2} \boldsymbol{\theta}_{S_c}^{\top}\boldsymbol{K}_{S_c, S_c} \boldsymbol{\theta}_{S_c}-\boldsymbol{\theta}_{S_c}^{\top} \boldsymbol{K}_{S_c, \boldsymbol{q}}
\end{equation}

For EC-NNK-Means, we set $\lambda=0$ for the last two epochs of training and during inference. 
Therefore, the OOD scores for EC-NNK-Means and C-EC-NNK-Means are computed using \eqref{eq:reconstruction-error} and  \eqref{eq:c-error}, respectively, but using a dictionary that was learned under entropy constraints.

\section{OOD Detection Experiments}
\label{sec:experiments}

\subsection{Datasets}

We used three datasets to empirically measure OOD detection performance: \textbf{20 Newsgroups} \cite{LANG1995331}, \textbf{Banking77} \cite{casanueva-etal-2020-efficient}, and \textbf{CLINC150} \cite{larson-etal-2019-evaluation}. For 20 Newsgroups and Banking77, we randomly selected 25\%, 50\%, and 75\% of the classes to form the ID training set $D_\text{ID}$, following \citet{zhang2021deep}. The remaining classes were used as OOD data at test time. 
CLINC150 contains a designated OOD label, and the rest of the dataset was used as $D_\text{ID}$ following \citet{lin-gu-2023-flats}.
We also report results on the larger \textbf{AG News} \cite{Zhang2015CharacterlevelCN} in \autoref{app:Additional Results}. Dataset statistics, splits, and other details can be found in \autoref{app:Dataset Details}.
\begin{table*}[ht!]
\centering
\small
\renewcommand{\arraystretch}{0.8}
\begin{tabular}{llrrrrrrc}
\toprule
    && \multicolumn{3}{c}{\textbf{20 Newsgroups}} & \multicolumn{3}{c}{\textbf{Banking77}} & \multirow{2}{*}{\textbf{CLINIC-150}} \\
    \cmidrule(r){3-6} \cmidrule(lr){6-8} 
    & \% \textbf{ID Classes} $\rightarrow$& \multicolumn{1}{c}{\bf 25\%} & \multicolumn{1}{c}{\bf 50\%} & \multicolumn{1}{c}{\bf 75\%} & \multicolumn{1}{c}{\bf 25\%} & \multicolumn{1}{c}{\bf 50\%} & \multicolumn{1}{c}{\bf 75\%} &  \\
    \midrule
    \multirow{4}{*}{\rotatebox{90}{Label-Blind}} 
    & KNN & 78.50 & 81.40 & 82.28 & 93.21 & 92.34 & 92.29 & 97.89 \\
    & kMeans & 78.75 & \textbf{81.94} & \textbf{83.19} & 93.21 & 92.34 & 92.35 & 97.89 \\ \cmidrule[0.03em](lr){2-9}
    & NNK-Means$^\dagger$ & \textbf{79.24} & 81.00 & 82.23 & \textbf{93.38} & 92.68 & \textbf{92.56} & 98.16 \\ 
    & EC-NNK-Means$^\dagger$ & 79.07 & 80.76 & 81.86 & 93.27 & \textbf{92.72} & 92.30 & \textbf{98.23} \\ 
    \midrule
    \multirow{8}{*}{\rotatebox{90}{Label-Aware }}
    & MSP & 72.80 & 79.52 & 81.84 & 88.00 & 88.26 & 89.63 & 96.43 \\ 
    & Energy & 72.58 & 80.35 & \textbf{83.94} & 88.34 & 88.95 & 90.27 & 97.07 \\ 
    & D2U & 73.38 & 80.54 & 83.69 & 87.84 & 89.10 & 90.24 & 97.15 \\
    & BLOOD & 66.11 & 71.86 & 69.75 & 73.93 & 70.09 & 69.72 & 87.12 \\
    & Mahalanobis & 75.61 & 73.16 & 75.92 & 93.17 & 92.63 & \textbf{92.78} & 97.81 \\
    & C-kMeans & 78.64 & 82.01 & 82.99 & 93.02 & 92.21 & 92.25 & 97.90 \\ \cmidrule[0.03em](lr){2-9}
    & C-NNK-Means$^\dagger$ & \textbf{79.30} & 81.96 & 83.06 & 93.32 & 92.62 & 92.69 & 97.97 \\ 
    & C-EC-NNK-Means$^\dagger$ & 79.12 & \textbf{82.45} & 83.21 & \textbf{93.48} & \textbf{92.73} & 92.75 & \textbf{98.03} \\
\bottomrule
\end{tabular}
\caption{\textbf{AUROC} for OOD detection on 3 datasets with fine-tuned Sentence-BERT representations. Label-aware methods incorporate ID labels during training, while label-blind methods are unable to do so. Results are averaged over 5 random seeds. The best ($\uparrow$) label-aware and label-blind methods in each column are \textbf{bolded}. NNK-Means and its variants, marked with $\dagger$, are our methods.}
\label{tab:AUROC-Results}
\end{table*}

\subsection{Baselines and Models}

We compared \textbf{NNK-Means}, our extended \textbf{EC-NNK-Means}, and their respective class-wise versions, \textbf{C-NNK-Means} and \textbf{C-EC-NNK-Means}, with 8 popular or recently proposed methods. For \textit{classifier-based} OOD detection methods, we chose \textbf{Maximum Softmax Probability (MSP)} \cite{hendrycks2017a}, \textbf{Energy} \cite{liu2020energy}, and \textbf{Distance-to-Uniform (D2U)} \cite{yilmaz2022d2u}. For \textit{distance-based} OOD detection methods, we evaluated \textbf{Mahalanobis} \cite{Lee-etal-mahalanobis} and \textbf{KNN} \cite{Sun2022ICMLDeepNN}. We also compare against \textbf{BLOOD} \cite{jelenic2023out}, which leverages between-layer representations, as well as \textbf{kMeans} and its class-wise version \textbf{C-kMeans}. For better illustration, we reclassified these methods into \textbf{Label-Aware} and \textbf{Label-Blind} methods, as shown in \autoref{tab:AUROC-Results}. Label-Aware methods incorporate ID labels during training, while Label-Blind methods do not. Details of each method are provided in the \autoref{app:Baselines Details}.

We used Sentence-BERT \cite{DBLP:journals/corr/abs-1908-10084} (82M parameters) as the encoder \(E\). Implementation details can be found in \autoref{app:implementation}. \autoref{app:Hyperparameters Tuning} details our hyper-parameter tuning process for some OOD detection methods.

\subsection{Evaluation Metrics}

We treat OOD detection as a binary classification task, where the OOD class is considered the positive sample. Following \citet{hendrycks2017a} and \citet{podolskiy2021revisiting}, we used standard evaluation metrics \textbf{AUROC}, \textbf{AUPR}, and \textbf{FPR@95}. We also used \textbf{Inference Time} (in seconds) as an additional metric to account for the efficiency of the OOD detection methods. 
\autoref{app:evaluation metrics} provides more details.

\section{OOD Detection Results and Analysis}
\label{sec:results}

Table \ref{tab:AUROC-Results} shows the AUROC of the baselines and our proposed methods on the three evaluation datasets. AUPR and FPR@95 results are in \autoref{app:Additional Results}.

\paragraph{NNK-Means outperforms baselines} 
Overall, we find that NNK-Means and its variants have better performance than all baselines in most cases (71\% of experimental settings\footnote{In 5 out the 7 settings in \autoref{tab:AUROC-Results}, NNK-Means and its variants have the highest AUROC}).
Furthermore, classifier-based approaches tend to perform worse than clustering and distance-based ones. 
Classifier-based approaches only had the best performance in one of the tested settings, and consistently achieved the low AUROC in all others. 
Despite their benefits with regards to efficiency, these approaches do not provide competitive performance.
\begin{figure}[!ht]
    \centering
    \includegraphics[width=\linewidth]{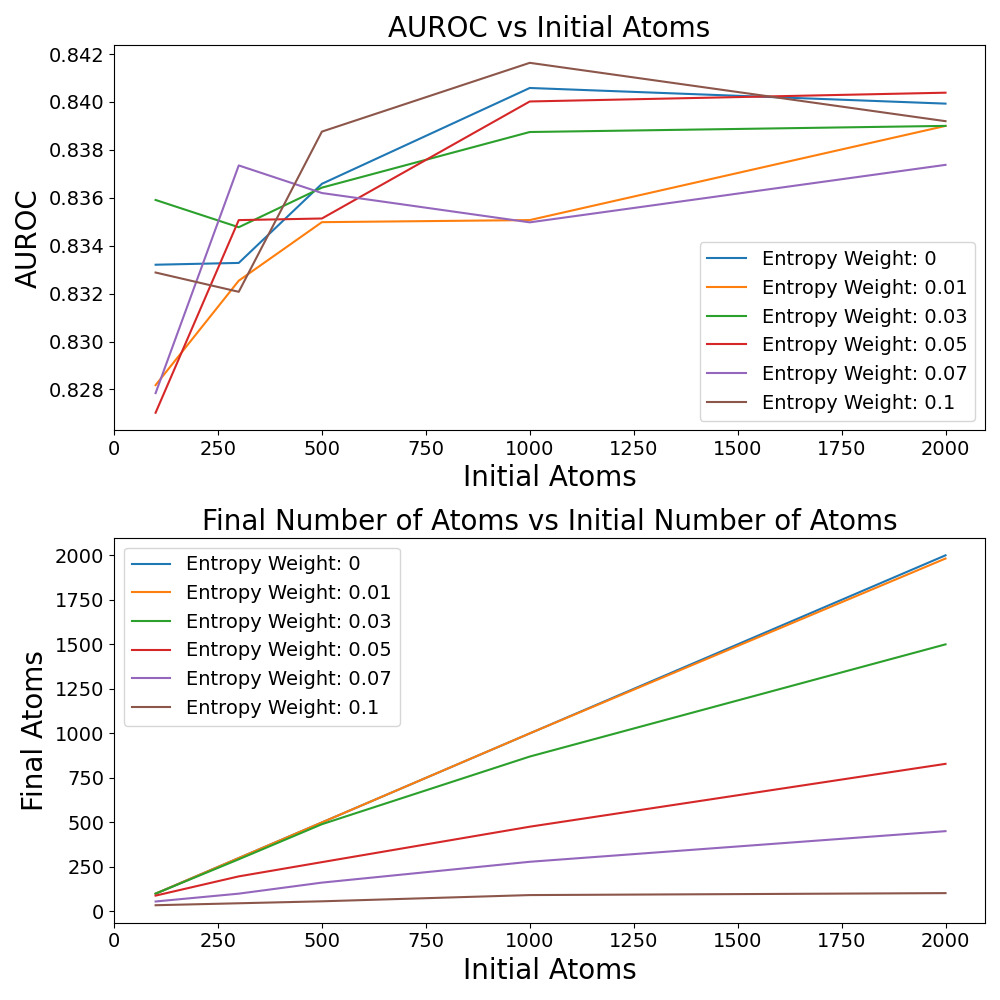}
    \caption{Final number of atoms and AUROC for different values of Entropy Constraint hyper-parameter $\lambda$, and number of starting atoms. Reported on 20 Newsgroups with 25\% ID classes. EC-NNK-Means can yield competitive performance with 90\% less memory usage. }
    \label{fig:ablation}
\end{figure}

\paragraph{NNK-Means effectively leverages ID labels} 
NNK-Means and kMeans are the only methods that are applicable when no labelled ID data is present, but can also incorporate label information if it is available.
Nonetheless, we find that NNK-Means is better able to leverage ID labels when compared to kMeans.
The label-aware variants of NNK-Means performed better than their label-blind counterparts in 86\% of cases.
In contrast, kMeans outperformed C-kMeans in 57\% of settings. 
Therefore, although kMeans can incorporate ID labels, NNK-Means uses this information more  effectively.

\paragraph{NNK-Means has low storage requirements}
An advantage of clustering-based methods is that the storage requirement depends on the number of clusters, not the size of the dataset. 
NNK-Means performs better than all baselines while only storing 2K cluster centers instead of all 15K instances from CLINIC-150. This is 87\% less storage than the best of our baselines, KNN.

\begin{table}[tb]
\centering
\small
\renewcommand{\arraystretch}{0.9}
\resizebox{\columnwidth}{!}{
\begin{tabular}{llrrr}
\toprule
    && \textbf{20 NG} & \textbf{Banking} & \textbf{CLINIC} \\
    \midrule
    \multirow{4}{*}{\rotatebox{90}{Label-Blind}}
       & KNN & 1.41 & 1.68 & 7.01 \\ 
       & kMeans & 0.25 & 0.49 & 0.72 \\ \cmidrule[0.03em](lr){2-5}
       & NNK-Means$^\dagger$ & \textbf{0.23} & 0.44 & 0.60 \\ 
       & EC-NNK-Means$^\dagger$ & \textbf{0.23} & \textbf{0.40} & \textbf{0.59} \\ 
        \midrule
        \multirow{4}{*}{\rotatebox{90}{Label-Aware}}
       & Mahalanobis & \textbf{0.04} & \textbf{0.37} & \textbf{0.64} \\
       & C-kMeans & 2.27 & 15.79 & 79.32 \\ \cmidrule[0.03em](lr){2-5}
       & C-NNK-Means$^\dagger$ & 2.27 & 16.68 & 85.67 \\ 
       & C-EC-NNK-Means$^\dagger$ & 2.24 & 15.51 & 85.89 \\ 
\bottomrule
\end{tabular}
}

\caption{OOD detection \textbf{Inference Time} in seconds, measured on the test set and averaged over all runs for each dataset. The best ($\downarrow$) label-aware and label-blind methods in each column are \textbf{bolded}. We don't report this metric for MSP, Energy, D2U and BLOOD as explained in \autoref{app:evaluation metrics}. NNK-Means and its variants, marked with $\dagger$, are our methods.}
\label{tab:Time-Results}
\end{table}
\begin{figure}[tb]
    \centering
    \includegraphics[width=\linewidth]{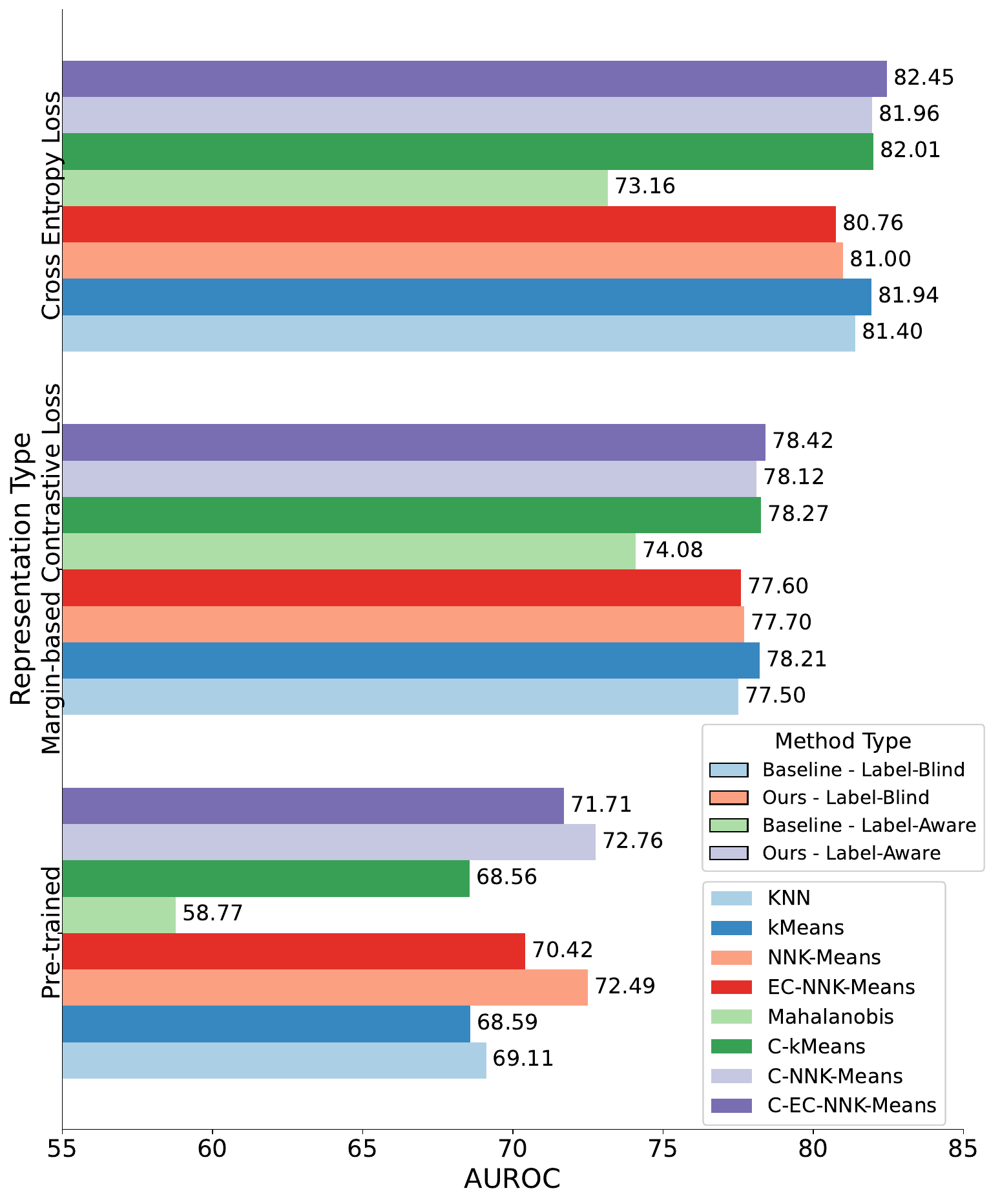}
    \caption{OOD Detection AUROC on 20 Newsgroups with 50\% ID classes, with different Sentence-BERT embeddings. Results are averaged over 5 random seeds.}
    \label{fig:different_embeddings}
\end{figure}
\begin{table*}[ht!]
    \centering
    \small
    \renewcommand{\arraystretch}{0.8}
    \resizebox{\linewidth}{!}{
    \begin{tabular}{p{13.5cm}llr}
    \toprule
         \textbf{20 NG Document} & \textbf{Label} & \textbf{OOD / ID }& \textbf{Error} \\
         \midrule
        Here it is Zoom 14.4k  FAX/DATA v.32bis modem.  I have evreything only purchased in January.  Will happily provide the Fax/Comm. software and BOX and manuals. I am selling this for ONLY \$125+s/h COD. \textbf{[Name]} \textbf{[Phone Number]} FEEL FREE TO CALL for quickest service. & \hlc[class_blue]{misc.forsale} & ID & 0.09 \\ \midrule
        
        NAPA remanufactured large 4 barrel carburetor for 78-80 big-block 360/440 Dodge.  Part \#4-244.  New in box w/manifold gasket. Retail: \$345.00 NAPA price: \$250.00 Your price \$100.00 + shipping
          &  \hlc[class_blue]{misc.forsale} & ID & 0.17 \\ \midrule
        If you'd like to find a home for that beekeeping equipment you'll never use again, here's a likely victim, uh, customer. To make a deal, call: \textbf{[Name]} \textbf{[Phone Number]} & \hlc[class_blue]{misc.forsale} & ID & 0.44 \\ \midrule
         I have several isolation amplifier boards that are the ideal interface for EEG and ECG.  Isolation is essential for safety when connecting line-powered equipment to electrodes on the body.  These boards incorporate the Burr-Brown 3656 isolation module that currently sells for \$133, plus other op amps to produce an overall voltage gain of 350-400.  They are like new and guaranteed good.  \$20 postpaid, schematic included.  Please email me for more data.
          & \hlc[class_ood]{sci.med} & OOD & 0.20 \\ \midrule 
          The title says it all.  Contact me via EMAIL if you would can help me out... \textbf{[Name]} University of Louisville P.S.  I KNOW IT IS DISCONTINUED.  I want someone who would like to sell an old copy. & \hlc[class_ood]{sci.electronics} & OOD & 0.24 \\ \midrule
          For all people that are interested in every aspect of the 2600 try the zine: 2600 connection \$1 cash to : \textbf{[Name]} \textbf{[Address]} for sample & \hlc[class_ood]{sci.electronics} & OOD & 0.16 \\
          \bottomrule
         
    \end{tabular}
    }
    
    \caption{
    Example of OOD instances overlapping with ID data from the visualization in \autoref{fig:nnkmviz}, with identical label colors.
    Last column represents the NNK-Means Error, as presented in \eqref{eq:reconstruction-error}.
    All ID and OOD instances mention the purchase or sale of a product, despite belonging to different classes. \textbf{Bolded} text is edited from the original to preserve anonymity.
    }
    \label{tab:manual}
\end{table*}

\autoref{fig:ablation} shows how the proposed entropy constraint can reduce storage requirements even further.
When working with EC-NNK-Means, the goal is to start with a large initial dictionary size and choose successively larger values of entropy-constraint hyperparameter $\lambda$ until the final dictionary is of the desired size.
We find that with $\lambda=0.1$, less than 100 atoms remain in the final dictionary, but the OOD detection AUROC is comparable or better than a dictionary with 2K atoms and $\lambda=0$.
Therefore, we show that EC-NNK-Means can achieve \textbf{comparable or better performance} than NNK-Means and KNN while using \textbf{95\% and 97\% less memory}, respectively. 
This reduced memory requirement is particularly useful when working with large datasets - where storing and running computations on the entire ID train set may be challenging.

\paragraph{NNK-Means has reduced inference time}
\autoref{tab:Time-Results} shows that NNK-Means is significantly faster than KNN as operating on the smaller, learned dictionaries is quicker than working with the entire ID train dataset. 
In particular, on the CLINIC-150 dataset, EC-NNK-Means provides an $11 \times$ reduction in inference time relative to KNN. 
Class-wise variants of NNK-Means have higher inference time because they involve iterating through one dictionary per ID class, an operation that is not parallelized like the computations in NNK-Means. 

\begin{figure}[ht]
    \centering
    \includegraphics[width=\columnwidth]{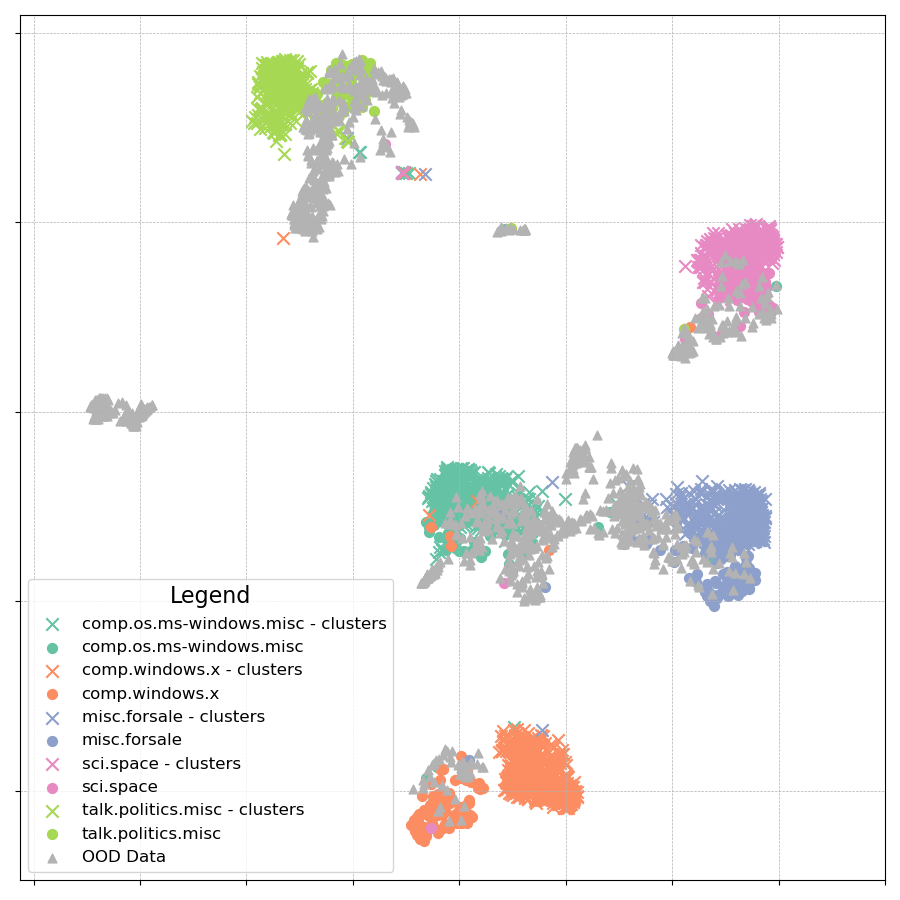}
    \caption{2D visualization of 20 Newsgroups validation dataset and learned clusters, with 25\% ID classes.}
    \label{fig:nnkmviz}
\end{figure}

\paragraph{Competitive performance with different embeddings}
A key benefit of NNK-Means is its applicability in various settings, independent of the embeddings being used. 
To empirically validate the performance of our methods when using different representations, we evaluate OOD detection performance using two different types of embeddings, as presented in \autoref{tab:pt_margin}.
We report results on the 20 Newsgroups dataset, comparing pre-trained embeddings and embeddings from a Sentence-BERT model fine-tuned with margin-based contrastive loss, as in \citet{zhou-etal-2021-contrastive}.

We find that NNK-Means provides competitive performance, outperforming all baselines even when different representations are used (see \autoref{fig:different_embeddings}).
In particular, when using pre-trained representations, NNK-Means performs significantly better than all other baselines (4 AUROC points better than best baseline, KNN). \autoref{app:Additional Results} provides further results with different types of embeddings.

\paragraph{Qualitative analysis of clustering}
\autoref{fig:nnkmviz} uses UMAP \citep{mcinnes2020umap} to visualize the results of our clustering process. We find that our clustering works as expected: when a dictionary learned on the training set is used to cluster the validation data, instances with the same class label are assigned to the same clusters.
We also see separate clusters of OOD data when their class labels are substantially different from the ID labels.
In some cases, there is overlap between OOD instances and ID data, such as the blue ``misc.forsale'' class. Analysing the text of these OOD documents shows that this overlap occurs because the OOD and ID instances are similar 
(see \autoref{tab:manual}).

\section{Conclusion}
\label{sec:conclusion}
We address the problem of OOD detection using NNK-Means, a soft-clustering algorithm. 
NNK-Means achieves state-of-the-art performance across 4 benchmark datasets, while requiring lower storage and improving computational efficiency relative to previous approaches that perform comparably. 
We introduce EC-NNK-Means, an extension of NNK-Means, and show that it can lead to further improvements in efficiency while matching or improving OOD detection performance. 
Our methods provide competitive performance regardless of the availability of labels or the type of embeddings used, and yield intuitive clustering of input data.
Future work will explore applying our algorithms to analyze large pretrained  datasets.

\clearpage

\section*{Ethical Considerations}

Our work aims to enable the robust and reliable deployment of Language Models by appropriately flagging OOD data and preventing inaccurate or unpredictable output. We do not anticipate any risks or harmful consequences stemming from our work. Our code and models will be publicly released in the future to ensure our work is reproducible. All datasets used in this paper are publicly available. 

\section*{Limitations}
There is a multitude of approaches for OOD detection, however, we were only able to compare against a subset of these approaches. 
Furthermore, our datasets, models, and experiments are all English-only. 
Finally, our experiments used data from classes that were unseen during training to simulate OOD data. 
In practice, there are many different ways a system may encounter OOD instances, and our experiments may not have covered them all. 

\section*{Acknowledgements}
This research was funded in part by the USC-Amazon Center on Secure and Trustworthy Machine Learning and by the National Science Foundation under grant CCF-2009032.

\bibliography{anthology,custom}

\begin{thebibliography}{47}
\expandafter\ifx\csname natexlab\endcsname\relax\def\natexlab#1{#1}\fi

\bibitem[{Breunig et~al.(2000)Breunig, Kriegel, Ng, and Sander}]{10.1145/335191.335388}
Markus~M. Breunig, Hans-Peter Kriegel, Raymond~T. Ng, and J\"{o}rg Sander. 2000.
\newblock \href {https://doi.org/10.1145/335191.335388} {Lof: identifying density-based local outliers}.
\newblock \emph{SIGMOD Rec.}, 29(2):93–104.

\bibitem[{Casanueva et~al.(2020)Casanueva, Tem{\v{c}}inas, Gerz, Henderson, and Vuli{\'c}}]{casanueva-etal-2020-efficient}
I{\~n}igo Casanueva, Tadas Tem{\v{c}}inas, Daniela Gerz, Matthew Henderson, and Ivan Vuli{\'c}. 2020.
\newblock \href {https://doi.org/10.18653/v1/2020.nlp4convai-1.5} {Efficient intent detection with dual sentence encoders}.
\newblock In \emph{Proceedings of the 2nd Workshop on Natural Language Processing for Conversational AI}, pages 38--45, Online. Association for Computational Linguistics.

\bibitem[{Chen et~al.(2022)Chen, Bi, Gao, and Sun}]{chen2022holistic}
Sishuo Chen, Xiaohan Bi, Rundong Gao, and Xu~Sun. 2022.
\newblock Holistic sentence embeddings for better out-of-distribution detection.
\newblock \emph{arXiv preprint arXiv:2210.07485}.

\bibitem[{Chen et~al.(2023)Chen, Yang, Bi, and Sun}]{chen2023fine}
Sishuo Chen, Wenkai Yang, Xiaohan Bi, and Xu~Sun. 2023.
\newblock Fine-tuning deteriorates general textual out-of-distribution detection by distorting task-agnostic features.
\newblock \emph{arXiv preprint arXiv:2301.12715}.

\bibitem[{Elsahar and Gall{\'e}(2019)}]{elsahar-galle-2019-annotate}
Hady Elsahar and Matthias Gall{\'e}. 2019.
\newblock \href {https://doi.org/10.18653/v1/D19-1222} {To annotate or not? predicting performance drop under domain shift}.
\newblock In \emph{Proceedings of the 2019 Conference on Empirical Methods in Natural Language Processing and the 9th International Joint Conference on Natural Language Processing (EMNLP-IJCNLP)}, pages 2163--2173, Hong Kong, China. Association for Computational Linguistics.

\bibitem[{Engan et~al.(1999)Engan, Aase, and Husoy}]{engan1999method}
Kjersti Engan, Sven~Ole Aase, and J~Hakon Husoy. 1999.
\newblock Method of optimal directions for frame design.
\newblock In \emph{1999 IEEE International Conference on Acoustics, Speech, and Signal Processing. Proceedings. ICASSP99 (Cat. No. 99CH36258)}, volume~5, pages 2443--2446. IEEE.

\bibitem[{He et~al.(2003)He, Xu, and Deng}]{he2003discovering}
Zengyou He, Xiaofei Xu, and Shengchun Deng. 2003.
\newblock Discovering cluster-based local outliers.
\newblock \emph{Pattern recognition letters}, 24(9-10):1641--1650.

\bibitem[{Hendrycks and Gimpel(2017)}]{hendrycks2017a}
Dan Hendrycks and Kevin Gimpel. 2017.
\newblock \href {https://openreview.net/forum?id=Hkg4TI9xl} {A baseline for detecting misclassified and out-of-distribution examples in neural networks}.
\newblock In \emph{International Conference on Learning Representations}.

\bibitem[{Jeleni{\'c} et~al.(2023)Jeleni{\'c}, Juki{\'c}, Tutek, Puljiz, and {\v{S}}najder}]{jelenic2023out}
Fran Jeleni{\'c}, Josip Juki{\'c}, Martin Tutek, Mate Puljiz, and Jan {\v{S}}najder. 2023.
\newblock Out-of-distribution detection by leveraging between-layer transformation smoothness.
\newblock \emph{arXiv preprint arXiv:2310.02832}.

\bibitem[{Kim et~al.(2023)Kim, Jung, Na, Jang, Park, and Choi}]{kim2023pseudo}
Jaeyoung Kim, Kyuheon Jung, Dongbin Na, Sion Jang, Eunbin Park, and Sungchul Choi. 2023.
\newblock Pseudo outlier exposure for out-of-distribution detection using pretrained transformers.
\newblock \emph{arXiv preprint arXiv:2307.09455}.

\bibitem[{Kriegel et~al.(2009)Kriegel, Kr\"{o}ger, Schubert, and Zimek}]{10.1007/978-3-642-01307-2_86}
Hans-Peter Kriegel, Peer Kr\"{o}ger, Erich Schubert, and Arthur Zimek. 2009.
\newblock \href {https://doi.org/10.1007/978-3-642-01307-2_86} {Outlier detection in axis-parallel subspaces of high dimensional data}.
\newblock In \emph{Proceedings of the 13th Pacific-Asia Conference on Advances in Knowledge Discovery and Data Mining}, PAKDD '09, page 831–838, Berlin, Heidelberg. Springer-Verlag.

\bibitem[{Lang et~al.(2022)Lang, Zheng, Sun, Huang, Si, and Li}]{lang2022estimating}
Hao Lang, Yinhe Zheng, Jian Sun, Fei Huang, Luo Si, and Yongbin Li. 2022.
\newblock Estimating soft labels for out-of-domain intent detection.
\newblock \emph{arXiv preprint arXiv:2211.05561}.

\bibitem[{Lang(1995)}]{LANG1995331}
Ken Lang. 1995.
\newblock \href {https://doi.org/https://doi.org/10.1016/B978-1-55860-377-6.50048-7} {Newsweeder: Learning to filter netnews}.
\newblock In Armand Prieditis and Stuart Russell, editors, \emph{Machine Learning Proceedings 1995}, pages 331--339. Morgan Kaufmann, San Francisco (CA).

\bibitem[{Larson et~al.(2019)Larson, Mahendran, Peper, Clarke, Lee, Hill, Kummerfeld, Leach, Laurenzano, Tang, and Mars}]{larson-etal-2019-evaluation}
Stefan Larson, Anish Mahendran, Joseph~J. Peper, Christopher Clarke, Andrew Lee, Parker Hill, Jonathan~K. Kummerfeld, Kevin Leach, Michael~A. Laurenzano, Lingjia Tang, and Jason Mars. 2019.
\newblock \href {https://doi.org/10.18653/v1/D19-1131} {An evaluation dataset for intent classification and out-of-scope prediction}.
\newblock In \emph{Proceedings of the 2019 Conference on Empirical Methods in Natural Language Processing and the 9th International Joint Conference on Natural Language Processing (EMNLP-IJCNLP)}, pages 1311--1316, Hong Kong, China. Association for Computational Linguistics.

\bibitem[{Lee et~al.(2018)Lee, Lee, Lee, and Shin}]{Lee-etal-mahalanobis}
Kimin Lee, Kibok Lee, Honglak Lee, and Jinwoo Shin. 2018.
\newblock \href {https://proceedings.neurips.cc/paper_files/paper/2018/file/abdeb6f575ac5c6676b747bca8d09cc2-Paper.pdf} {A simple unified framework for detecting out-of-distribution samples and adversarial attacks}.
\newblock In \emph{Advances in Neural Information Processing Systems}, volume~31. Curran Associates, Inc.

\bibitem[{Lewis et~al.(2021)Lewis, Stenetorp, and Riedel}]{lewis-etal-2021-question}
Patrick Lewis, Pontus Stenetorp, and Sebastian Riedel. 2021.
\newblock \href {https://doi.org/10.18653/v1/2021.eacl-main.86} {Question and answer test-train overlap in open-domain question answering datasets}.
\newblock In \emph{Proceedings of the 16th Conference of the European Chapter of the Association for Computational Linguistics: Main Volume}, pages 1000--1008, Online. Association for Computational Linguistics.

\bibitem[{Liang et~al.(2018)Liang, Li, and Srikant}]{liang2018enhancing}
Shiyu Liang, Yixuan Li, and R.~Srikant. 2018.
\newblock \href {https://openreview.net/forum?id=H1VGkIxRZ} {Enhancing the reliability of out-of-distribution image detection in neural networks}.
\newblock In \emph{International Conference on Learning Representations}.

\bibitem[{Liang et~al.(2017)Liang, Li, and Srikant}]{liang2017enhancing}
Shiyu Liang, Yixuan Li, and Rayadurgam Srikant. 2017.
\newblock Enhancing the reliability of out-of-distribution image detection in neural networks.
\newblock \emph{arXiv preprint arXiv:1706.02690}.

\bibitem[{Lin and Gu(2023)}]{lin-gu-2023-flats}
Haowei Lin and Yuntian Gu. 2023.
\newblock \href {https://doi.org/10.18653/v1/2023.emnlp-main.554} {{FL}at{S}: Principled out-of-distribution detection with feature-based likelihood ratio score}.
\newblock In \emph{Proceedings of the 2023 Conference on Empirical Methods in Natural Language Processing}, pages 8956--8963, Singapore. Association for Computational Linguistics.

\bibitem[{Liu et~al.(2024)Liu, Zhan, Lu, Feng, Xue, and Wu}]{liu-etal-2024-good-llms}
Bo~Liu, Li-Ming Zhan, Zexin Lu, Yujie Feng, Lei Xue, and Xiao-Ming Wu. 2024.
\newblock \href {https://aclanthology.org/2024.lrec-main.720} {How good are {LLM}s at out-of-distribution detection?}
\newblock In \emph{Proceedings of the 2024 Joint International Conference on Computational Linguistics, Language Resources and Evaluation (LREC-COLING 2024)}, pages 8211--8222, Torino, Italia. ELRA and ICCL.

\bibitem[{Liu et~al.(2022)Liu, Lewis, Riedel, and Stenetorp}]{liu-etal-2022-challenges}
Linqing Liu, Patrick Lewis, Sebastian Riedel, and Pontus Stenetorp. 2022.
\newblock \href {https://doi.org/10.18653/v1/2022.findings-naacl.155} {Challenges in generalization in open domain question answering}.
\newblock In \emph{Findings of the Association for Computational Linguistics: NAACL 2022}, pages 2014--2029, Seattle, United States. Association for Computational Linguistics.

\bibitem[{Liu et~al.(2020)Liu, Wang, Owens, and Li}]{liu2020energy}
Weitang Liu, Xiaoyun Wang, John Owens, and Yixuan Li. 2020.
\newblock Energy-based out-of-distribution detection.
\newblock \emph{Advances in neural information processing systems}, 33:21464--21475.

\bibitem[{Loshchilov and Hutter(2017)}]{loshchilov2017decoupled}
Ilya Loshchilov and Frank Hutter. 2017.
\newblock Decoupled weight decay regularization.
\newblock \emph{arXiv preprint arXiv:1711.05101}.

\bibitem[{Manolache et~al.(2021)Manolache, Brad, and Burceanu}]{manolache2021date}
Andrei Manolache, Florin Brad, and Elena Burceanu. 2021.
\newblock Date: Detecting anomalies in text via self-supervision of transformers.
\newblock \emph{arXiv preprint arXiv:2104.05591}.

\bibitem[{McInnes et~al.(2020)McInnes, Healy, and Melville}]{mcinnes2020umap}
Leland McInnes, John Healy, and James Melville. 2020.
\newblock \href {http://arxiv.org/abs/1802.03426} {Umap: Uniform manifold approximation and projection for dimension reduction}.

\bibitem[{Ouyang et~al.(2023)Ouyang, Cao, Gao, Wu, Zhang, and Dai}]{ouyang2023prefix}
Yawen Ouyang, Yongchang Cao, Yuan Gao, Zhen Wu, Jianbing Zhang, and Xinyu Dai. 2023.
\newblock On prefix-tuning for lightweight out-of-distribution detection.
\newblock In \emph{Proceedings of the 61st Annual Meeting of the Association for Computational Linguistics (Volume 1: Long Papers)}, pages 1533--1545.

\bibitem[{Plank(2016)}]{plank2016nonstandard}
Barbara Plank. 2016.
\newblock \href {https://arxiv.org/abs/1608.07836} {What to do about non-standard (or non-canonical) language in nlp}.

\bibitem[{Podolskiy et~al.(2021)Podolskiy, Lipin, Bout, Artemova, and Piontkovskaya}]{podolskiy2021revisiting}
Alexander Podolskiy, Dmitry Lipin, Andrey Bout, Ekaterina Artemova, and Irina Piontkovskaya. 2021.
\newblock Revisiting mahalanobis distance for transformer-based out-of-domain detection.
\newblock In \emph{Proceedings of the AAAI Conference on Artificial Intelligence}, pages 13675--13682.

\bibitem[{Reimers and Gurevych(2019)}]{DBLP:journals/corr/abs-1908-10084}
Nils Reimers and Iryna Gurevych. 2019.
\newblock \href {http://arxiv.org/abs/1908.10084} {Sentence-bert: Sentence embeddings using siamese bert-networks}.
\newblock \emph{CoRR}, abs/1908.10084.

\bibitem[{Salehi et~al.(2022)Salehi, Mirzaei, Hendrycks, Li, Rohban, and Sabokrou}]{salehi2022a}
Mohammadreza Salehi, Hossein Mirzaei, Dan Hendrycks, Yixuan Li, Mohammad~Hossein Rohban, and Mohammad Sabokrou. 2022.
\newblock \href {https://openreview.net/forum?id=aRtjVZvbpK} {A unified survey on anomaly, novelty, open-set, and out of-distribution detection: Solutions and future challenges}.
\newblock \emph{Transactions on Machine Learning Research}.

\bibitem[{Shekkizhar and Ortega(2020)}]{shekkizhar2020graph}
Sarath Shekkizhar and Antonio Ortega. 2020.
\newblock Graph construction from data by non-negative kernel regression.
\newblock In \emph{ICASSP 2020-2020 IEEE International Conference on Acoustics, Speech and Signal Processing (ICASSP)}, pages 3892--3896. IEEE.

\bibitem[{Shekkizhar and Ortega(2022)}]{shekkizhar2022nnk}
Sarath Shekkizhar and Antonio Ortega. 2022.
\newblock \href {https://doi.org/10.23919/EUSIPCO55093.2022.9909928} {{NNK-Means: Data summarization using dictionary learning with non-negative kernel regression}}.
\newblock In \emph{2022 30th European Signal Processing Conference (EUSIPCO)}, pages 2161--2165. IEEE.

\bibitem[{Shu et~al.(2021)Shu, Benajiba, Mansour, and Zhang}]{shu2021odist}
Lei Shu, Yassine Benajiba, Saab Mansour, and Yi~Zhang. 2021.
\newblock Odist: Open world classification via distributionally shifted instances.
\newblock In \emph{Findings of the Association for Computational Linguistics: EMNLP 2021}, pages 3751--3756.

\bibitem[{Sun et~al.(2022)Sun, Ming, Zhu, and Li}]{Sun2022ICMLDeepNN}
Yiyou Sun, Yifei Ming, Xiaojin Zhu, and Yixuan Li. 2022.
\newblock \href {https://proceedings.mlr.press/v162/sun22d.html} {Out-of-distribution detection with deep nearest neighbors}.
\newblock In \emph{Proceedings of the 39th International Conference on Machine Learning}, volume 162 of \emph{Proceedings of Machine Learning Research}, pages 20827--20840. PMLR.

\bibitem[{Wu et~al.(2022)Wu, He, Yan, Gao, Zeng, Zheng, Zhao, Jiang, Wu, and Xu}]{wu2022revisit}
Yanan Wu, Keqing He, Yuanmeng Yan, QiXiang Gao, Zhiyuan Zeng, Fujia Zheng, Lulu Zhao, Huixing Jiang, Wei Wu, and Weiran Xu. 2022.
\newblock Revisit overconfidence for ood detection: Reassigned contrastive learning with adaptive class-dependent threshold.
\newblock In \emph{Proceedings of the 2022 Conference of the North American Chapter of the Association for Computational Linguistics: Human Language Technologies}, pages 4165--4179.

\bibitem[{Xu et~al.(2022)Xu, Ren, and Jia}]{xu2022contrastive}
Albert Xu, Xiang Ren, and Robin Jia. 2022.
\newblock Contrastive novelty-augmented learning: Anticipating outliers with large language models.
\newblock \emph{arXiv preprint arXiv:2211.15718}.

\bibitem[{Xu et~al.(2020)Xu, He, Yan, Liu, Liu, and Xu}]{xu2020deep}
Hong Xu, Keqing He, Yuanmeng Yan, Sihong Liu, Zijun Liu, and Weiran Xu. 2020.
\newblock A deep generative distance-based classifier for out-of-domain detection with mahalanobis space.
\newblock In \emph{Proceedings of the 28th International Conference on Computational Linguistics}, pages 1452--1460.

\bibitem[{Xu et~al.(2021)Xu, Ren, Zhang, Feng, and Xiong}]{xu2021unsupervised}
Keyang Xu, Tongzheng Ren, Shikun Zhang, Yihao Feng, and Caiming Xiong. 2021.
\newblock Unsupervised out-of-domain detection via pre-trained transformers.
\newblock \emph{arXiv preprint arXiv:2106.00948}.

\bibitem[{Yang et~al.(2023)Yang, Song, Ren, Lyu, Wang, Zhuo, Liu, Wang, Foster, and Zhang}]{yang-etal-2023-distribution}
Linyi Yang, Yaoxian Song, Xuan Ren, Chenyang Lyu, Yidong Wang, Jingming Zhuo, Lingqiao Liu, Jindong Wang, Jennifer Foster, and Yue Zhang. 2023.
\newblock \href {https://doi.org/10.18653/v1/2023.emnlp-main.276} {Out-of-distribution generalization in natural language processing: Past, present, and future}.
\newblock In \emph{Proceedings of the 2023 Conference on Empirical Methods in Natural Language Processing}, pages 4533--4559, Singapore. Association for Computational Linguistics.

\bibitem[{Yilmaz and Toraman(2022)}]{yilmaz2022d2u}
Eyup Yilmaz and Cagri Toraman. 2022.
\newblock D2u: Distance-to-uniform learning for out-of-scope detection.
\newblock In \emph{Proceedings of the 2022 Conference of the North American Chapter of the Association for Computational Linguistics: Human Language Technologies}, pages 2093--2108.

\bibitem[{Zeng et~al.(2021)Zeng, He, Yan, Xu, and Xu}]{zeng2021adversarial}
Zhiyuan Zeng, Keqing He, Yuanmeng Yan, Hong Xu, and Weiran Xu. 2021.
\newblock Adversarial self-supervised learning for out-of-domain detection.
\newblock In \emph{Proceedings of the 2021 Conference of the North American Chapter of the Association for Computational Linguistics: Human Language Technologies}, pages 5631--5639.

\bibitem[{Zhan et~al.(2021)Zhan, Liang, Liu, Fan, Wu, and Lam}]{zhan2021out}
Li-Ming Zhan, Haowen Liang, Bo~Liu, Lu~Fan, Xiao-Ming Wu, and Albert Lam. 2021.
\newblock Out-of-scope intent detection with self-supervision and discriminative training.
\newblock \emph{arXiv preprint arXiv:2106.08616}.

\bibitem[{Zhang et~al.(2021)Zhang, Xu, and Lin}]{zhang2021deep}
Hanlei Zhang, Hua Xu, and Ting-En Lin. 2021.
\newblock Deep open intent classification with adaptive decision boundary.
\newblock In \emph{Proceedings of the AAAI Conference on Artificial Intelligence}, pages 14374--14382.

\bibitem[{Zhang et~al.(2015)Zhang, Zhao, and LeCun}]{Zhang2015CharacterlevelCN}
Xiang Zhang, Junbo~Jake Zhao, and Yann LeCun. 2015.
\newblock Character-level convolutional networks for text classification.
\newblock In \emph{NIPS}.

\bibitem[{Zhou et~al.(2021{\natexlab{a}})Zhou, Liu, and Chen}]{zhou2021contrastive}
Wenxuan Zhou, Fangyu Liu, and Muhao Chen. 2021{\natexlab{a}}.
\newblock Contrastive out-of-distribution detection for pretrained transformers.
\newblock \emph{arXiv preprint arXiv:2104.08812}.

\bibitem[{Zhou et~al.(2021{\natexlab{b}})Zhou, Liu, and Chen}]{zhou-etal-2021-contrastive}
Wenxuan Zhou, Fangyu Liu, and Muhao Chen. 2021{\natexlab{b}}.
\newblock \href {https://doi.org/10.18653/v1/2021.emnlp-main.84} {Contrastive out-of-distribution detection for pretrained transformers}.
\newblock In \emph{Proceedings of the 2021 Conference on Empirical Methods in Natural Language Processing}, pages 1100--1111, Online and Punta Cana, Dominican Republic. Association for Computational Linguistics.

\bibitem[{Zhou et~al.(2022)Zhou, Liu, and Qiu}]{zhou2022knn}
Yunhua Zhou, Peiju Liu, and Xipeng Qiu. 2022.
\newblock Knn-contrastive learning for out-of-domain intent classification.
\newblock In \emph{Proceedings of the 60th Annual Meeting of the Association for Computational Linguistics (Volume 1: Long Papers)}, pages 5129--5141.

\end{thebibliography}

\appendix
\section{Additional Results}
\label{app:Additional Results}

We provide additional results on AUPR (see \autoref{tab:AUPR-Results}) and FPR@95 (see \autoref{tab:FPR@95-Results}) for the 3 main datasets aligned with \autoref{tab:AUROC-Results}. To demonstrate that our methods perform relatively better on larger datasets, we also include results on AG News; see \autoref{tab:agnews} for more details.
Additionally, to show the competitive performance of our proposed methods with different representations (detailed analysis in Section \ref{sec:results}), we also provide the AUROC results with \textbf{Pre-Trained} Embeddings and \textbf{Margin-based Contrastive Loss} Embeddings (see \autoref{tab:pt_margin}), which are reported for 50\% ID classes ratio using label-blind and label-aware methods on 20 Newsgroups.

\begin{table*}[ht!]
\centering
\small
\renewcommand{\arraystretch}{0.8}
\begin{tabular}{llrrrrrrc}
\toprule
    && \multicolumn{3}{c}{\textbf{20 Newsgroups}} & \multicolumn{3}{c}{\textbf{Banking77}} & \multirow{2}{*}{\textbf{CLINIC-150}} \\
    \cmidrule(r){3-6} \cmidrule(lr){6-8} 
    & \% \textbf{ID Classes} $\rightarrow$& \multicolumn{1}{c}{\bf 25\%} & \multicolumn{1}{c}{\bf 50\%} & \multicolumn{1}{c}{\bf 75\%} & \multicolumn{1}{c}{\bf 25\%} & \multicolumn{1}{c}{\bf 50\%} & \multicolumn{1}{c}{\bf 75\%} &  \\
    \midrule
    \multirow{4}{*}{\rotatebox{90}{Label-Blind}} 
    & KNN & 50.26 & 77.40 & 90.93 & 85.72 & 92.27 & 97.04 & 99.47 \\
    & kMeans & 50.72 & 77.91 & \textbf{91.53} & \textbf{85.73} & 92.27 & 97.09 & 99.47 \\ \cmidrule[0.03em](lr){2-9}
    & NNK-Means$^\dagger$ & \textbf{51.11} & \textbf{76.88} & 90.88 & 85.69 & 92.68 & \textbf{97.22}& \textbf{99.53}\\ 
    & EC-NNK-Means$^\dagger$ & 50.75 & 76.76 &90.79 & 85.63 & \textbf{92.72}& 97.07& \textbf{99.53} \\ 
    \midrule
    \multirow{8}{*}{\rotatebox{90}{Label-Aware}}
    & MSP & \textbf{53.66} & \textbf{82.41} & \textbf{93.44} & 72.08 & 87.73 & 95.70 & 99.01 \\ 
    & Energy & 47.42 & 81.98 & 94.21 & 72.60 & 88.17 & 95.95 & 99.19 \\ 
    & D2U & 51.16 & 82.43 & 94.08 & 70.42 & 88.36 & 95.92 & 99.21 \\
    & BLOOD & 39.69 & 72.87 & 86.08 & 48.86 & 85.65 & 95.65 & 96.52 \\
    & Mahalanobis & 47.72 & 70.61 & 88.98 & 85.23 & 92.50 & \textbf{97.27} & 99.44 \\
    & C-kMeans & 50.21 & 77.80 & 91.30 & 85.05 & 91.83 & 97.00 & 99.45 \\ \cmidrule[0.03em](lr){2-9}
    & C-NNK-Means$^\dagger$ & 51.26& 77.71 & 91.21 & \textbf{85.73} &92.53 & 97.26& 99.49 \\ 
    & C-EC-NNK-Means$^\dagger$ & 51.11& 77.82& 91.19 & 86.12 &\textbf{92.74} & 97.25& \textbf{99.50}\\
\bottomrule
\end{tabular}
\caption{\textbf{AUPR} for OOD detection on 3 datasets with fine-tuned Sentence-BERT representations. Label-aware methods incorporate ID labels during training, while label-blind methods are unable to do so. Results are averaged over 5 random seeds. The best ($\uparrow$) label-aware and label-blind methods in each column are \textbf{bolded}. NNK-Means and its variants, marked with $\dagger$, are our methods.}
\label{tab:AUPR-Results}
\end{table*}
\begin{table*}[ht!]
\centering
\small
\renewcommand{\arraystretch}{0.8}
\begin{tabular}{llrrrrrrc}
\toprule
    && \multicolumn{3}{c}{\textbf{20 Newsgroups}} & \multicolumn{3}{c}{\textbf{Banking77}} & \multirow{2}{*}{\textbf{CLINIC-150}} \\
    \cmidrule(r){3-6} \cmidrule(lr){6-8} 
    & \% \textbf{ID Classes} $\rightarrow$& \multicolumn{1}{c}{\bf 25\%} & \multicolumn{1}{c}{\bf 50\%} & \multicolumn{1}{c}{\bf 75\%} & \multicolumn{1}{c}{\bf 25\%} & \multicolumn{1}{c}{\bf 50\%} & \multicolumn{1}{c}{\bf 75\%} &  \\
    \midrule
    \multirow{4}{*}{\rotatebox{90}{Label-Blind}} 
    & KNN & 71.36 & 75.25 & 72.42 & 33.99 & 33.69 & \textbf{37.65} & 10.90 \\
    & kMeans & \textbf{69.65} & \textbf{71.12}& \textbf{68.59} & 33.39 & 33.77 & 37.95 & 10.78 \\ \cmidrule[0.03em](lr){2-9}
    & NNK-Means$^\dagger$ & 70.50 & 75.43 & 74.61 &\textbf{31.98} &\textbf{33.28} &37.75 & 8.74 \\ 
    & EC-NNK-Means$^\dagger$ & 71.03 & 75.45 & 76.25 &33.08 & 33.31& 39.32& \textbf{8.60} \\ 
    \midrule
    \multirow{8}{*}{\rotatebox{90}{Label-Aware}}
    & MSP & 87.13 & 85.18 & 82.92 & 50.41 & 54.47 & 55.92 & 17.00 \\ 
    & Energy & 85.74 & 83.58 & 81.17 & 44.55 & 46.82 & 46.27 & 12.90 \\ 
    & D2U & 85.44 & 84.00 & 81.05 & 45.21 & 45.95 & 46.47 & 12.70 \\
    & BLOOD & 91.50 & 86.37 & 90.15 & 77.46 & 80.88 & 83.22 & 58.38 \\
    & Mahalanobis & 77.37 & 86.09 & 87.19 & 34.40 & 33.83 & 36.33 & \textbf{9.60} \\
    & C-kMeans & 70.54 & 72.06 & 69.80 & 34.17 & 34.23 & 37.62 & 10.88 \\ \cmidrule[0.03em](lr){2-9}
    & C-NNK-Means$^\dagger$ & \textbf{70.20} & 72.45 & \textbf{69.52}& \textbf{32.43}&32.65 & 36.25 & 10.04 \\ 
    & C-EC-NNK-Means$^\dagger$ & 70.98 & \textbf{70.59} & 69.65&33.64 &\textbf{32.44} & \textbf{36.15} & 10.00 \\
\bottomrule
\end{tabular}
\caption{\textbf{FPR@95} for OOD detection on 3 datasets with fine-tuned Sentence-BERT representations. Label-aware methods incorporate ID labels during training, while label-blind methods are unable to do so. Results are averaged over 5 random seeds. The best ($\downarrow$) label-aware and label-blind methods in each column are \textbf{bolded}. NNK-Means and its variants, marked with $\dagger$, are our methods.}
\label{tab:FPR@95-Results}
\end{table*}
\begin{table*}[ht!]
\centering
\small
\renewcommand{\arraystretch}{0.8}
\begin{tabular}{llrrrrrrc}
\toprule
    && \multicolumn{2}{c}{\textbf{AUROC} ($\uparrow$)} & \multicolumn{2}{c}{\textbf{AUPR} ($\uparrow$)} & \multicolumn{2}{c}{\textbf{FPR@95} ($\downarrow$)} & \multirow{2}{*}{\textbf{Infer. Time ($\downarrow$)}} \\
    \cmidrule(r){3-4} \cmidrule(lr){5-6} \cmidrule(lr){7-8} 
    & \% \textbf{ID Classes} $\rightarrow$& \multicolumn{1}{c}{\bf 50\%} & \multicolumn{1}{c}{\bf 75\%} & \multicolumn{1}{c}{\bf 50\%} & \multicolumn{1}{c}{\bf 75\%} & \multicolumn{1}{c}{\bf 50\%} & \multicolumn{1}{c}{\bf 75\%} &  \\
    \midrule
    \multirow{4}{*}{\rotatebox{90}{Label-Blind}} 
    & KNN & 83.75 & 93.09 & 83.03 & 97.23 & 46.47 & 31.71 & 18.50 \\ 
    & kMeans & 83.54 & 93.49 & 82.74 & 97.30 & 47.35 & 27.61 & \textbf{0.89} \\ \cmidrule[0.03em](lr){2-9}
    & NNK-Means$^\dagger$ & 83.91 & 93.22 & 82.70 & 97.30 & 45.33 & 30.45 & 1.44 \\ 
    & EC-NNK-Means$^\dagger$ & 84.07 & 93.43 & 83.64 & 97.31 & 46.01 & 28.36 & 0.95 \\ 
    \midrule
    \multirow{8}{*}{\rotatebox{90}{Label-Aware}}
    & MSP & 82.84 & 86.07 & 83.97 & 94.78 & 53.58 & 55.80 & - \\ 
    & Energy & 79.90 & 86.68 & 79.01 & 94.81 & 55.13 & 46.68 & - \\ 
    & D2U & 82.84 & 87.72 & 84.06 & 95.25 & 53.56 & 46.68 & - \\ 
    & BLOOD & 77.95 & 86.16 & 75.35 & 93.73 & 53.62 & 51.45 & - \\
    & Mahalanobis & 83.42 & 92.10 & 83.79 & 96.79 & 53.54 & 34.08 & \textbf{0.02} \\
    & C-kMeans & 83.72 & 93.42 & 82.96 & 97.25 & 47.59 & 27.74 & 2.03 \\ \cmidrule[0.03em](lr){2-9}
    & C-NNK-Means$^\dagger$ & 83.26 & 93.37 & 82.20 & 97.31 & 46.92 & 28.72 & 2.18 \\ 
    & C-EC-NNK-Means$^\dagger$ & \textbf{86.30} & \textbf{94.47} & \textbf{86.47} & \textbf{97.98} & \textbf{45.87} & \textbf{26.80} & 1.98 \\
\bottomrule
\end{tabular}
\caption{OOD detection performance on \textbf{AG News} are reported for \textbf{AUROC}, \textbf{AUPR}, \textbf{FPR@95} and \textbf{Inference Time} in seconds with fine-tuned Sentence-BERT representations. Label-aware methods incorporate ID labels during training, while label-blind methods are unable to do so. Results are averaged over 5 random seeds. The best label-aware and label-blind methods in each column are \textbf{bolded}. We do not report Inferece Time for MSP, Energy, D2U, and BLOOD as discussed in \autoref{app:evaluation metrics}. NNK-Means and its variants, marked with $\dagger$, are our methods.}
\label{tab:agnews}
\end{table*}
\begin{table*}[ht]
\centering
\small
\renewcommand{\arraystretch}{0.8}
\begin{tabular}{llrrrrrc}
\toprule
    && \multicolumn{3}{c}{\textbf{Pre-Trained}} & \multicolumn{3}{c}{\textbf{Margin-based Contrastive Loss}} \\
    \cmidrule(r){3-5} \cmidrule(lr){6-8} 
    & \% \textbf{ID Classes} $\rightarrow$& \multicolumn{1}{c}{\bf 25\%} & \multicolumn{1}{c}{\bf 50\%} & \multicolumn{1}{c}{\bf 75\%} & \multicolumn{1}{c}{\bf 25\%} & \multicolumn{1}{c}{\bf 50\%} & \multicolumn{1}{c}{\bf 75\%}  \\
    \midrule
    \multirow{4}{*}{\rotatebox{90}{Label-Blind}} 
    & KNN & 71.75 & 69.11 & 67.63 & 79.95 & 77.50 & 78.82 \\ 
    & kMeans & 70.35 & 68.59 & 67.43 & \textbf{80.32} & \textbf{78.21} & \textbf{80.07} \\ \cmidrule[0.03em](lr){2-8}
    & NNK-Means$^\dagger$ & \textbf{75.52} & \textbf{72.49} & \textbf{69.52} & 80.27 & 77.70 & 79.10 \\ 
    & EC-NNK-Means$^\dagger$ & 75.01 & 70.42 & 67.82 & 80.15 & 77.60 & 79.01 \\ 
    \midrule
    \multirow{8}{*}{\rotatebox{90}{Label-Aware}}
    & MSP & - & - & - & 78.26 & 75.59 & 77.36 \\
    & Energy & - & - & - & 78.55 & 75.46 & 78.42 \\
    & D2U & - & - & - & 79.19 & 76.37 & 78.77 \\
    & BLOOD & - & - & - & 69.02 & 73.64 & 65.74 \\
    & Mahalanobis & 62.75 & 58.77 & 57.64 & 76.07 & 74.08 & 72.91 \\
    & C-kMeans & 70.29 & 68.56 & 68.45 & 80.29 & 78.27 & \textbf{79.73} \\ \cmidrule[0.03em](lr){2-8}
    & C-NNK-Means$^\dagger$ & 75.08 & \textbf{72.76} & \textbf{70.48} & 80.42 & 78.12 & 79.67 \\ 
    & C-EC-NNK-Means$^\dagger$ & \textbf{76.49} & 71.71 & 70.20 & \textbf{80.53} & \textbf{78.42} & 79.55 \\
\bottomrule
\end{tabular}
\caption{\textbf{AUROC} comparison with \textbf{Pre-Trained} Embeddings and \textbf{Margin-based Contrastive Loss} Embeddings. Results are reported for 50\% ID classes ratio using label-blind and label-aware methods on 20 Newsgroups. The best ($\uparrow$) label-aware and label-blind methods in each column are \textbf{bolded}. We do not report Pre-Trained Embedding results for MSP, Energy, D2U, and BLOOD as discussed in \autoref{app:Baselines Details}. NNK-Means and its variants, marked with $\dagger$, are our methods.}
\label{tab:pt_margin}
\end{table*}

\section{Datasets}
\label{app:Dataset Details}

In this section, we specifically introduce the four datasets we used and how they were partitioned. Each dataset was divided into training/validation/test sets. In \autoref{tab:datasets}, we provide the statistical details of these datasets before distinguishing between ID and OOD classes.

\begin{table}[ht]
\centering
\small
\resizebox{0.48\textwidth}{!}{
\begin{tabular}{lcccc}
\toprule
    \textbf{Dataset} & \textbf{\# Training} & \textbf{\# Validation} & \textbf{\# Test} & \textbf{\# Classes} \\
    \midrule
    20 Newsgroups & 15076 & 1885 & 1885 & 20 \\
    Banking77 & 9002 & 1001 & 3080 & 77 \\
    CLINC150 & 15000 & 3100 & 5500 & 150+1\\
    AG News & 112800 & 7200 & 7600 & 4 \\
\bottomrule
\end{tabular}
}
\caption{Dataset summary with statistical details about the training, validation, and test sets along with the number of classes. Note that the number of training examples is initial.}
\label{tab:datasets}
\end{table}

\paragraph{20 Newsgroups \cite{LANG1995331}} 20 Newsgroups is a widely used benchmark for text classification, consisting of approximately 18000 newsgroup documents organized into 20 classes, each representing a specific topic such as politics, religion, science, and technology. We utilized the 20 Newsgroups dataset provided by \texttt{scikit-learn} and removed headers, signature blocks, and quotation blocks respectively as suggested. Following \citet{zhou-etal-2021-contrastive}, we divided the whole dataset into training/validation/test sets in an 80/10/10 ratio using stratified sampling based on class labels. For the training set, we randomly selected 25\%, 50\%, and 75\% of the classes as ID classes and removed the remaining classes, resulting in the dataset $D_{IN}$. In the validation and test sets, these selected classes were considered as IN class during the OOD detection phase, while the other classes were treated as OOD class.

\paragraph{Banking77 \cite{casanueva-etal-2020-efficient}} Banking77 is a specialized dataset for intent classification in the banking domain. It consists of 13083 customer service queries categorized into 77 distinct classes, each representing a specific banking-related intent. We used the HuggingFace version of this dataset, which includes 10003 user queries in the training set and 3080 queries in the test set. We split its training set into training and validation sets in a 90/10 ratio and applied the same preprocessing steps to the training set as we did with the 20 Newsgroups.

\paragraph{CLINC150 \cite{larson-etal-2019-evaluation}} CLINC150 is a dataset tailored for OOD intent detection. It includes 150 distinct intent classes from various domains and one designated OOD class for evaluation. The dataset consists of a total of 22500 ID queries and 1200 OOD queries. We used the ID training data directly as our training set and combined the ID validation and test data with the OOD validation and test data to form our validation and test sets, respectively.

\paragraph{AG News \cite{Zhang2015CharacterlevelCN}} AG News is a topic classification dataset collected from various news sources, encompassing a total of four topics. We used the HuggingFace version of this dataset, which includes 120000 entries in the training set and 7600 entries in the test set. We extracted 6\% of the training data to form a validation set. When selecting 25\% of the classes as ID classes, AG News only includes one class, making it unsuitable for classification tasks. Therefore, we only used the 50\% and 75\% settings for our experiments. The rest of the processing is similar to that applied to the 20 Newsgroups dataset.

\section{Baselines and Models}
\label{app:Baselines Details}

In this section, we provide a more detailed introduction to our baselines. Mathematical notations follow the conventions established in Section \ref{sec:preliminaries}.

\paragraph{Maximum Softmax Probability (MSP)}
\citet{hendrycks2017a} propose this method as the baseline for detecting OOD examples which has been widely adopted. For MSP, $O(\boldsymbol{x}; E')$ is the maximum softmax probability among any of the classes:
\begin{equation}
    O(\boldsymbol{x}; E') = \max_{c \in \{1, \hdots, C\}} p_c(E'(\boldsymbol{x}))
\end{equation}
where $p_c(\cdot)$ refers to the softmax probability for class $c$. Note that this method is applicable only when using fine-tuned encoder $E'$.

\paragraph{Energy}
\citet{liu2020energy} introduces the free energy function to detect OOD samples, which can replace the Softmax Confidence Score to avoid the overconfidence problem of the softmax function. The ID data tends to have low energy scores while OOD data tends to have high scores. The free energy function is formulated as follows:
\begin{equation}
    \text{Energy}(\boldsymbol{x}) = \sum_{i=1}^{C} e^{f_i(E'(\boldsymbol{x}))}
\end{equation}
where $f_i(\cdot)$ represents the output logits for the $i$-th class, and $C$ is the number of all classes. The score $O(\boldsymbol{x}; E')$ is then defined as the negative of the energy: 
\begin{equation}
    O(\boldsymbol{x}; E') = -\text{Energy}(\boldsymbol{x})
\end{equation}
Note that this method is also applicable only when using fine-tuned encoder $E'$.

\paragraph{Distance to Uniform (D2U)}
Based on the idea that output distributions of OOD samples get closer to the uniform distribution than that of ID samples, \citet{yilmaz2022d2u} introduces Distance-to-Uniform (D2U), which utilizes the shape of the entire output distribution and calculates its distance to the uniform distribution as a metric to evaluate the likelihood of an example being OOD:
\begin{equation}
    O(\boldsymbol{x}; E') = \text{dst}(\boldsymbol{p}(E'(\boldsymbol{x})), U)
\end{equation}
where $\boldsymbol{p}(\cdot)$ is the output softmax distribution and $U$ refers to the uniform distribution. We follow \citet{yilmaz2022d2u}'s setting to use the KL divergence as the distance function. Note that this method is also applicable only when using fine-tuned encoder $E'$.

\paragraph{BLOOD} 
The BLOOD score proposed by \citet{jelenic2023out} is a method for detecting OOD data in Transformer-based models by examining the smoothness of transformations between intermediate layers. It utilizes the tendency of between-layer representation transformations of ID data to be smoother than the corresponding transformations of OOD data.
The smoothness of the transformation between layers $l$ and $l+1$ for an input $\boldsymbol{x}$ is quantified using the Frobenius norm of the Jacobian matrix for $l=1,\hdots, L-1$. This is given by:
\begin{equation}
    \phi_l(\boldsymbol{x}) = \| \boldsymbol{J}_l(\boldsymbol{h}_l) \|_F^2 = \sum_{i=1}^{d_{l+1}} \sum_{j=1}^{d_l} \left( \frac{\partial (f_{l+1})_i}{\partial (h_l)_j} \right)^2
\end{equation}
where $\boldsymbol{J}_l(\boldsymbol{h}_l)$ is the Jacobian matrix of the transformation from layer $l$ to $l+1$, $\boldsymbol{h}_l$ is the representation at layer $l$, and $\boldsymbol{f}_l: \mathbb{R}^{d_{l-1}} \rightarrow \mathbb{R}^{d_{l}}$ is the intermediate network layers, while $\boldsymbol{f}_L$ corresponds to the last layer, mapping to a vector of logits.
To reduce computational complexity, in practice, BLOOD uses an unbiased estimator of the smoothness measure with $r$ pairs of random vectors $\boldsymbol{v}_{l} \sim \mathcal{N}(\boldsymbol{0}_n, \boldsymbol{I}_{n})$ and $\boldsymbol{w}_{l} \sim \mathcal{N}(\boldsymbol{0}_m, \boldsymbol{I}_{m})$:
\begin{equation}
    \hat{\phi}_l(\boldsymbol{x}) = \frac{1}{r} \sum_{i=1}^r \left( \boldsymbol{w}_{l,i}^\top \boldsymbol{J}_l(\boldsymbol{h}_l) \boldsymbol{v}_{l,i} \right)^2
\end{equation}
The final BLOOD score for an input $\boldsymbol{x}$ can be computed as either the average smoothness score across all layers:
\begin{equation}
    \text{BLOOD}_M = \frac{1}{L-1} \sum_{l=1}^{L-1} \hat{\phi}_l(\boldsymbol{x})
\end{equation}
or the smoothness score at the last layer:
\begin{equation}
    \text{BLOOD}_L = \hat{\phi}_{L-1}(\boldsymbol{x})
\end{equation}
We follow \citet{jelenic2023out} to use $\text{BLOOD}_L$ as the uncertainty score of an instance $\boldsymbol{x}$ for its higher performance. Finally, the OOD score is defined as:
\begin{equation}
    O(\boldsymbol{x}; E') = -\text{BLOOD}_L
\end{equation}
Note that this method is also applicable only when using fine-tuned encoder $E'$.

\paragraph{Mahalanobis}
The Mahalanobis distance detector proposed by \citet{Lee-etal-mahalanobis} is a widely used OOD detection method that calculates the OOD score $O(\boldsymbol{x}; E)$ based on the distance of a test sample to the nearest ID class in the embedding space determined by $M$. It can be formulated as:
\begin{multline}
O(\boldsymbol{x}; E) = \min_{c \in \{1, \hdots, C\}} (E(\boldsymbol{x}) - \boldsymbol{\mu}_c)^\top \\
\boldsymbol{\Sigma}^{-1} (E(\boldsymbol{x}) - \boldsymbol{\mu}_c)
\end{multline}
where $\boldsymbol{\mu}_c$ is the mean of all of the representations of the instances in class $c$ and $\boldsymbol{\Sigma}$ is the covariance matrix. $\boldsymbol{\mu}_c$ and $\boldsymbol{\Sigma}$ can be estimated by:

\begin{align}
\hat{\boldsymbol{\mu}_c} &= \frac{1}{N_c} \sum_{\boldsymbol{x} \in D_{IN}^c} E(\boldsymbol{x}) \\
\hat{\boldsymbol{\Sigma}} &= \begin{aligned}[t]
&\frac{1}{N} \sum_{c \in \{1, \hdots, C\}} \sum_{\boldsymbol{x} \in D_{IN}^c} \\
& \quad (E(\boldsymbol{x}) - \boldsymbol{\mu}_c) (E(\boldsymbol{x}) - \boldsymbol{\mu}_c)^\top 
\end{aligned}
\end{align}
where $D_{IN}^c = \{\boldsymbol{x} \mid (\boldsymbol{x}, y) \in D_{IN}, y = c\}$ represents for the training data belonging to the class $c$, $N$ denotes the size of the training set, and $N_c$ is the number of training data belonging to the class $c$.

\paragraph{$k$-Nearest Neighbors (KNN)}
\citet{Sun2022ICMLDeepNN} investigate the effectiveness of using non-parametric nearest-neighbor distances for OOD detection on visual OOD detection benchmarks. We applied this approach to text data, where \( O(\boldsymbol{x}; E) \) represents the distance from the test sample to its  $k$-th nearest ID training sample in the normalized feature space. In our experiments, we set $k = 1$.

\paragraph{kMeans \& C-kMeans}
We also compare our approaches to the standard kMeans algorithm and its class-wise variant, C-kMeans, similar to the C-NNK-Means. In both cases, we use the reconstruction error as the OOD score \( O(\boldsymbol{x}; E) \). The number of clusters is a hyper-parameter, and their selection will be discussed in \autoref{app:Hyperparameters Tuning}.

\section{Implementation Details}
\label{app:implementation}
\paragraph{Fine-tuning} We fine-tuned the PLM for classification on the ID dataset and used the \texttt{all-distilroberta-v1} checkpoint from HuggingFace. In all cases, we used mean-pooling on token representations from the penultimate layer to generate sentence-level representations. We used 5 different random seeds and reported the average results to limit the effect of randomness for each setting.
All models were optimized with Cross Entropy Loss and \texttt{AdamW} \cite{loshchilov2017decoupled} as the optimizer, using a weight decay rate of 0.01 and a learning rate of $1 \times 10^{-5}$, with a linear learning rate decay. We used a batch size of 4 and fine-tuned the model for 5 epochs.

\paragraph{OOD Detection} After extracting embeddings, we ran our baselines and proposed methods on a single NVIDIA Tesla V100 GPU to ensure consistent measurement of inference time. 
We tuned hyper-parameters based on the validation set and reported the final results on the test set of each dataset. \autoref{app:Hyperparameters Tuning} provides more details of our hyper-parameter tuning.

\section{Evaluation Metrics}
\label{app:evaluation metrics}

Here, we introduce 3 standard metrics for OOD detection and the Inference Time in seconds we used to compare the complexity:

\paragraph{AUROC} The Area Under the Receiver Operating Characteristic Curve, plots the True Positive Rate (TPR) against the False Positive Rate (FPR) at various thresholds. A higher AUROC value indicates better performance.
\paragraph{AUPR} The Area Under the Precision-Recall Curve, evaluates the model’s precision and recall by plotting precision against recall for different thresholds. A higher AUPR value indicates better identification of OOD samples while maintaining high precision.
\paragraph{FPR@95} The False Positive Rate at 95\% True Positive Rate, measures the FPR when the TPR is fixed at 95\%. A lower FPR@95 value indicates fewer ID samples being misclassified as OOD, signifying a more reliable OOD detection model.
\paragraph{Inference Time} It serves as an additional metric to account for the complexity of the OOD detection methods.  We measured the time taken to obtain the OOD score of a given query $\boldsymbol{q}$ after extracting its representation from a PLM. Note that we do not report this for MSP, Energy, and D2U, as their inference involves minimally processing the logits, and so they have negligible inference time. We also do not report this for BLOOD since its inference process is significantly affected by the batch size. Additionally, BLOOD requires representations extracted from every layer of the model. So, despite doing limited processing after embeddings have been extracted, in practice, the complexity of this method is much higher than that of other classifier-based ones.

We provide the results of AUROC and Inference Time in Section \ref{sec:results}, and AUPR and FPR@95 results in \autoref{app:Additional Results}.

\section{Hyper-parameter Tuning}
\label{app:Hyperparameters Tuning}

KMeans, NNK-Means, and EC-NNK-Means select the number of dictionary atoms from $\{500, 1000, 2000, 4000\}$. For the class-wise versions, C-kMeans, C-NNK-Means, and C-EC-NNK-Means, due to the smaller size of each class compared to the overall dataset, the selection range is $\{50, 150, 250, 350\}$ instead. Additionally, for EC-NNK-Means and C-EC-NNK-Means, we also need to choose Entropy Constraint hyper-parameter $\lambda$ from $\{50, 150, 250, 350\}$. We tuned the hyper-parameters on the validation set of each dataset, selecting the optimal hyper-parameters based on AUROC for each dataset (and each known classes ratio), and obtained the final results on the test set. We applied the same hyper-parameter tuning process for the Pre-trained Embedding setting and the Margin-based Contrastive Loss Embedding setting. Detailed hyper-parameter choices for each setting can be found in \autoref{tab:HP1} and \autoref{tab:HP2}.

\begin{table*}[ht]
\centering
\small
\renewcommand{\arraystretch}{0.8}
\resizebox{\textwidth}{!}{
\begin{tabular}{llccccccccc}
\toprule
    && \multicolumn{3}{c}{\textbf{20 Newsgroups}} & \multicolumn{3}{c}{\textbf{Banking77}} & \multicolumn{2}{c}{\textbf{AG News}} & \multirow{2}{*}{\textbf{CLINC150}} \\
    \cmidrule(r){3-5} \cmidrule(lr){6-8} \cmidrule(lr){9-10}
    & \% \textbf{ID Classes} $\rightarrow$& \multicolumn{1}{c}{\bf 25\%} & \multicolumn{1}{c}{\bf 50\%} & \multicolumn{1}{c}{\bf 75\%} & \multicolumn{1}{c}{\bf 25\%} & \multicolumn{1}{c}{\bf 50\%} & \multicolumn{1}{c}{\bf 75\%} & \multicolumn{1}{c}{\bf 50\%} & \multicolumn{1}{c}{\bf 75\%} &  \\
    \midrule
    \multirow{6}{*}{\rotatebox{90}{Methods}} 
    & kMeans & 1000 & 500 & 500 & 2000 & 4000 & 1000 & 4000 & 500 & 2000 \\ 
    & C-kMeans & 250 & 50 & 50 & 32 & 32 & 32 & 350 & 50 & 50 \\ 
    & NNK-Means & 2000 & 2000 & 4000 & 1000 & 2000 & 4000 & 4000 & 4000 & 2000 \\ 
    & C-NNK-Means & 350 & 350 & 350 & 32 & 32 & 32 & 350 & 50 & 25 \\ 
    & EC-NNK-Means & (2000, 0.03) & (2000, 0.03) & (2000, 0.03) & (2000, 0.03) & (4000, 0.01) & (2000, 0.03) & (4000, 0.03) & (500, 0.01) & (2000, 0.05) \\ 
    & C-EC-NNK-Means & (350, 0.01) & (350, 0.07) & (350, 0.03) & (32, 0.01) & (32, 0.01) & (32, 0.01) & (50, 0.07) & (150, 0.07) & (50, 0.01) \\ 
\bottomrule
\end{tabular}
}
\caption{Hyper-parameter settings for different methods with Cross Entropy Loss Embeddings on 4 datasets. This is used for our main results in Section \ref{sec:results}. For EC-NNK-Means and C-EC-NNK-Means, the hyper-parameters are in the format of $(M, \lambda)$ where $M$ is the initial number of dictionary atoms and $\lambda$ is the hyper-parameter that controls the influence of entropy-constrained term, while others are only using $M$.}
\label{tab:HP1}
\end{table*}
\begin{table*}[ht]
\centering
\small
\renewcommand{\arraystretch}{0.8}
\begin{tabular}{llcccccc}
\toprule
    && \multicolumn{3}{c}{\textbf{Pre-trained}} & \multicolumn{3}{c}{\textbf{Margin-based Contrastive Loss}} \\
    \cmidrule(r){3-5} \cmidrule(lr){6-8} 
    & \% \textbf{ID Classes} $\rightarrow$& \multicolumn{1}{c}{\bf 25\%} & \multicolumn{1}{c}{\bf 50\%} & \multicolumn{1}{c}{\bf 75\%} & \multicolumn{1}{c}{\bf 25\%} & \multicolumn{1}{c}{\bf 50\%} & \multicolumn{1}{c}{\bf 75\%} \\
    \midrule
    \multirow{6}{*}{\rotatebox{90}{Methods}} 
    & kMeans & 2000 & 4000 & 4000 & 500 & 1000 & 500 \\ 
    & C-kMeans & 350 & 350 & 350 & 50 & 50 & 50 \\ 
    & NNK-Means & 2000 & 4000 & 4000 & 2000 & 2000 & 4000 \\ 
    & C-NNK-Means & 350 & 350 & 350 & 350 & 350 & 350 \\ 
    & EC-NNK-Means & (2000, 0.03) & (2000, 0.03) & (2000, 0.03) & (1000, 0.03) & (1000, 0.01) & (4000, 0.03) \\ 
    & C-EC-NNK-Means & (350, 0.05) & (350, 0.05) & (350, 0.05) & (250, 0.05) & (350, 0.03) & (350, 0.03) \\ 
\bottomrule
\end{tabular}
\caption{Hyper-parameter settings for different methods with Pre-trained Embeddings and Margin-based Contrastive Loss Embeddings on 20 Newsgroup. This is used for our additional analysis in Section \ref{sec:results} to show the competitive performance of our methods with different embeddings. For EC-NNK-Means and C-EC-NNK-Means, the hyper-parameters are in the format of $(M, \lambda)$ where $M$ is the initial number of dictionary atoms and $\lambda$ is the hyper-parameter that controls the influence of entropy-constrained term, while others are only using $M$.}
\label{tab:HP2}
\end{table*}
\begin{table*}[ht!]
    \centering
    \small
    \resizebox{\linewidth}{!}{
    \begin{tabular}{p{12.5cm}lr}
    \toprule
         \textbf{Text} & \textbf{Label} & \textbf{NNK-Means Error} \\
         \midrule
         I know nothing about Sun's but replacing pieces of libraries,shared or not, is straight forward on RS/6000's (all releases) Extract the appropriate pierce with ar; rebind the .o; and replace with ar. See Info for details. & \hlc[class_orange]{ID: comp.windows.x} & 0.19 \\ \midrule
        This is incorrect.  Sun has made no such claim regarding Devguide, and as manager of the Devguide engineering group  I can state with authority that work on Devguide is continuing apace.  We had quite a strong show of interest from the Devguide user community at last week's Solaris Developer's Conference. Devguide is being advocated not only as a valuable future builder tool, but as an important bit of transition technology that will help sustain current customers and facilitate their migration to the COSE Desktop Environment. If you have specific questions about Devguide availability, etc., you can contact \textbf{[Name]}, our Devguide Product Marketing person, at \textbf{[Phone Number]}. & \hlc[class_orange]{ID: comp.windows.x} & 0.24 \\ \midrule
         I was wondering if anyone knew of an interface to od ( octal dump ), I assume it would be called xod.  Actually, any viewer for a core file will do. I looked at export ( @ mit ) in the index of /contrib, but didn't find anything relevant. & \hlc[class_orange]{ID: comp.windows.x} & 0.19 \\ \midrule
         libXaw3d, the 3D Athena widget set will greatly improve the "sculptured" look. In Linux, with its shared, jump-table libs, you don't even have to recompile or relink. you merely have to: ln -sf /lib/libXaw3d.so.3.0 /lib/libXaw.so.3 & \hlc[class_orange]{ID: comp.windows.x} & 0.14 \\ 
         \bottomrule    
    \end{tabular}
    }
    \caption{Example of data from an ID cluster from the visualization in \autoref{fig:nnkmviz}, with identical label colors. Last column represents the NNK-Means Error, as presented in \eqref{eq:reconstruction-error}. \textbf{Bolded} text is edited from the original to preserve anonymity.}
    \label{tab:my_label2}
\end{table*}

\end{document}